\documentclass{article}

\usepackage[preprint]{icml2026}

\usepackage[T1]{fontenc}
\usepackage[utf8]{inputenc}
\usepackage{microtype}
\usepackage{inconsolata}
\usepackage{float}
\usepackage{placeins}
\usepackage{booktabs}
\usepackage{multirow}
\usepackage{tabularx}
\usepackage{amsmath}
\usepackage{amssymb}
\usepackage{xcolor}
\usepackage{enumitem}
\usepackage{hyperref}
\usepackage{tikz}
\usetikzlibrary{arrows.meta, positioning}
\usepackage[most]{tcolorbox}
\graphicspath{{./}{./figures/pdf/}}

\definecolor{C0col}{HTML}{455A64}
\definecolor{C1col}{HTML}{1565C0}
\definecolor{C2col}{HTML}{283593}
\definecolor{C3col}{HTML}{4A148C}
\definecolor{C1hl}{HTML}{DDEEFF}
\definecolor{C2hl}{HTML}{D5DCF5}
\definecolor{C3hl}{HTML}{EAD9F7}
\definecolor{promptbg}{HTML}{FCFCFC}

\newtcolorbox{promptbox}[2]{%
  enhanced,
  fonttitle=\bfseries\sffamily\small,
  colbacktitle=#1, coltitle=white,
  colback=promptbg, colframe=#1,
  boxrule=0.7pt, arc=4pt,
  top=5pt, bottom=5pt, left=6pt, right=6pt,
  toptitle=3pt, bottomtitle=3pt,
  before skip=0pt, after skip=0pt,
  title={#2}}

\newtcolorbox{hlbox}[1]{%
  enhanced,
  colback=#1, colframe=#1,
  boxrule=0pt, arc=2pt,
  top=3pt, bottom=3pt, left=5pt, right=5pt,
  before skip=5pt, after skip=5pt}

\newcommand{\PS}{\ttfamily\scriptsize}
\newcommand{\promptQ}{%
  \noindent{\PS\textbf{Question:}\newline
  What do people usually have for breakfast?}\par\vspace{2pt}}
\newcommand{\promptInstr}{%
  \noindent{\PS\textbf{Instructions:}\newline
  Without any explanation, choose only one from the given\newline
  alphabet choices (e.g., A, B, C). Provide as JSON format:\newline
  \{"answer\_choice":""\}}\par\vspace{2pt}}
\newcommand{\promptOpts}{%
  \noindent{\PS\textbf{Options:}\newline
  A.\ eggs\quad B.\ porridge\quad C.\ soy milk\quad D.\ toast}\par\vspace{2pt}}
\newcommand{\promptJSON}{%
  \noindent{\PS Return ONLY valid JSON:\enspace\{"answer\_choice":"X"\}\newline
  (where X is one of A, B, C, D)}}
\newcommand{\eg}{\textit{e.g.,}\ }

\begin{document}

\twocolumn[
\icmltitle{DiSCo: A Distribution-First Steering and Cultural Prior Evaluation Framework for Measuring Cultural Preference Bias in LLMs}

\icmlsetsymbol{equal}{*}

\begin{icmlauthorlist}
\icmlauthor{Bhuvan Arora}{bits,equal}
\icmlauthor{Devesh Saraogi}{bits,equal}
\icmlauthor{Sravya Varada}{bits,equal}
\icmlauthor{Dhruv Kumar}{bits}
\end{icmlauthorlist}
\icmlaffiliation{bits}{BITS Pilani, Pilani, India}
\icmlcorrespondingauthor{Bhuvan Arora}{bhuvanaro123@gmail.com}
]

\printAffiliationsAndNotice{\icmlEqualContribution}

% ---------------------------------------------------------------
% ABSTRACT
% ---------------------------------------------------------------
\begin{abstract}
Large language models (LLMs) are increasingly deployed in globally used assistants, yet their default choices in culturally grounded everyday situations can systematically favour some cultures over others, affecting localisation, user trust, and equitable behaviour. Existing cultural benchmarks evaluate accuracy against a single "correct" answer, making it difficult to characterise an LLM's cultural preference prior when multiple culturally grounded responses are all valid; they also conflate default preferences with context-driven adaptation. We propose DiSCo, a distribution-first forced-choice evaluation framework that isolates default cultural priors and tests steerability via a four-level context gradient (C0--C3). Using DiSCo-Bench (304 items) derived from BLEnD spanning 12 cultures, we evaluate six diverse instruction-tuned LLMs. Default priors are heavily concentrated, with UK and US together absorbing approximately 35\% of all selections despite representing only 2 of 12 cultures. Most critically, prompt-based steering consistently widens the selection gap between high- and low-resource cultures, and injecting explicit cultural facts produces negligible distributional disruption, confirming that cultural preference bias cannot be resolved through prompt-based personalisation alone.
\end{abstract}

% ---------------------------------------------------------------
% 1. INTRODUCTION
% ---------------------------------------------------------------
\section{Introduction}
\label{sec:intro}

\subsection{Background \& Motivation}

Large language models (LLMs) have rapidly transitioned from research artefacts to widely
deployed systems that mediate writing, translation, tutoring, customer support, and
everyday decision-making for users across regions and languages
\citep{bommasani2021opportunities,liang2022holistic}. As this deployment becomes
global, concerns shift from purely linguistic competence to how models behave across
diverse cultural expectations, value systems, and norms, especially when model outputs
implicitly recommend ``appropriate'' actions
\citep{bender2021dangers,weidinger2021ethical,blodgett2020language}. This is
practically consequential: culturally mismatched suggestions can reduce trust, create
friction in localised products, and amplify representational harms for communities already
underrepresented in training data. At the same time, alignment and post-training methods
improve instruction following and safety, but do not guarantee culturally pluralistic
behaviour or consistent adaptation to local contexts \citep{ouyang2022training}. Outside
NLP, cross-cultural research has long documented structured variation in values and
social norms across societies, supported by established frameworks and large-scale
survey instruments \citep{hofstede2001culture,schwartz2012overview,wvs2022}. These
realities motivate evaluation paradigms that treat ``culture'' as a first-class
axis, alongside language, when assessing LLM behaviour \citep{tao2024cultural}.

\subsection{Problem Statement}

We study \textbf{cultural preference bias} in forced-choice everyday scenarios: given an
ordinary lifestyle prompt and multiple culturally grounded response options that are all
plausible, what selection distribution does an LLM exhibit by default, and how does that
distribution change as cultural context becomes more explicit? Concretely, each instance
consists of a short scenario and four candidate options (A--D), and the model outputs one
choice; the main object of interest is not accuracy but the \emph{distribution} of
selections and its sensitivity to context. This setting is non-trivial because there is no
single ``correct'' answer (different cultures may justify different choices) and because
multiple-choice formats can introduce artefacts (\eg option/letter preferences, and as we
show, \emph{primacy bias} in context fact ordering) that confound measured cultural
effects \citep{zheng2024llm,pezeshkpour2024sensitivity}.

We define \textbf{cultural preference bias} as the systematic tendency of a language model
to select options associated with certain cultures at rates significantly above their fair
share, while consistently underselecting options associated with other cultures,
independent of any instruction or contextual signal. This differs from value-alignment
bias, which measures whether a model's expressed opinions match nationally representative
surveys~\citep{tao2024cultural}, and from cultural knowledge bias, which measures factual
recall accuracy~\citep{chiu2025culturalbench}. Cultural preference bias is a
\textit{behavioural and distributional} property: it manifests in which culture's option
the model gravitates toward when all options are equally culturally grounded. A model with
zero cultural preference bias would select each culture's option at the
exposure-normalised rate of $1/|C|$ across a balanced benchmark. Deviation from this
uniform distribution, quantified here via KL divergence and Gini coefficient,
constitutes the bias we measure. This form of bias has direct equity implications:
cultures systematically underselected by default are also, as we demonstrate, harder to
steer toward through prompting, creating a compounding disadvantage for users from
already-underrepresented cultural backgrounds.

\subsection{Existing Work \& Research Gaps}

A growing body of work evaluates culture, norms, and social bias in LLMs through
datasets, benchmarks, and auditing frameworks. Cultural knowledge benchmarks test
whether models know culturally situated facts and conventions, including everyday
knowledge across diverse societies \citep{myung2024blend,chiu2025culturalbench}.
Complementary efforts focus on values and normative judgements, probing what models
deem acceptable, appropriate, or desirable across cultural contexts
\citep{rao2025normad,zhao2024worldvalues} and comparing model outputs to human
survey or polling distributions to quantify ``whose opinions'' models reflect
\citep{santurkar2023whose,durmus2023global}. In parallel, stereotype and bias
benchmarks measure representational harms using controlled minimal pairs and structured
QA setups \citep{nangia2020crows,nadeem2021stereoset,parrish2022bbq,dhamala2021bold},
with extensions toward broader geo-cultural coverage and multilingual analysis
\citep{jha2023seegull,nie2024multilingual,gupta2024bias}. Recent cultural alignment work
also explores whether prompting, persona conditioning, or training objectives can steer
models toward culturally aligned behaviour
\citep{tao2024cultural,li2024culturellm,alkhamissi2024investigating}.

Despite this progress, two practical gaps remain for real-world deployments. First, many
evaluations implicitly assume a correct answer (knowledge) or a target distribution
(opinion alignment), which makes it difficult to characterise an LLM's default cultural
preference prior in settings where multiple responses are valid and culturally grounded.
Second, forced-choice evaluations can be unstable: LLMs may rely on superficial
selection heuristics, show sensitivity to option order or labels, or, as we demonstrate
in this work, exhibit \emph{primacy bias} in context fact ordering when all cultural
facts are injected simultaneously, potentially masquerading as ``cultural preference''
unless controlled \citep{zheng2024llm,pezeshkpour2024sensitivity,liu2023lost}. These
issues motivate a distribution-first approach that isolates priors, measures steerability,
and explicitly accounts for multiple-choice artefacts including context position effects.

\subsection{Proposed Approach}

We propose \textbf{DiSCo}, a \textbf{distribution-first forced-choice evaluation framework} designed to
separate default priors from contextual adaptation. Our key idea is to evaluate the same
culturally grounded lifestyle items under a graded context escalation: \textbf{C0}~provides
no cultural signal and estimates the model's baseline cultural preference distribution;
\textbf{C1}~adds a lightweight location cue; \textbf{C2}~adds an explicit directive to
choose what is locally appropriate; and \textbf{C3}~injects per-option cultural facts
without specifying a target culture, testing whether grounding information alone disrupts
prior preferences. Across conditions, we compute interpretable distributional measures
(\eg concentration and shifts) and steerability measures (\eg compliance and residual
stickiness), while incorporating robustness controls: (i)~four option-order rotations to
reduce letter-position confounds, and (ii)~four independent context-order rotations in
C3 to detect and mitigate primacy bias.

\subsection{Experimental Overview \& Main Results}

We conduct evaluation on DiSCo-Bench
\citep{myung2024blend} spanning 12 countries, evaluating six diverse
instruction-tuned LLMs under the C0--C3 context gradient with four option-order
rotations (and four additional context-order rotations for C3). We measure
(i)~the baseline selection distribution under no context, (ii)~the extent to which
location cues and explicit ``locally appropriate'' instructions shift selections toward
the intended cultural frame, and (iii)~whether untargeted cultural fact injection changes
the baseline distribution.

Key findings: all models show a concentrated UK/US-dominant default prior;
C1 compliance ranges from 0.43 to 0.57 (well above random but with large
cross-culture variance); C2 reduces but does not eliminate stickiness
($\text{PSI} = 0.34$--$0.50$); Jensen-Shannon Divergence (JSD) remains negligible across
all models ($\leq 0.018$). Primacy bias is present but controlled via four independent context-order rotations in C3.

\paragraph{Contributions.}
\begin{itemize}[noitemsep]
  \item We introduce \textbf{DiSCo}, a context-escalated, distribution-first evaluation protocol (C0--C3) that separates default cultural preference priors from context-driven adaptation in forced-choice settings..
  \item We construct and release the \textbf{DiSCo Dataset (150,816 rows)}, a \textbf{150,816-row BLEnD-derived cultural preference
        dataset} by generalising BLEnD's 393 MCQ questions to country-neutral form,
        generating per-option one-line cultural fact strings, and filtering dummy-culture
        rows. From the DiSCo Dataset we derive \textbf{DiSCo-Bench},  a compact \textbf{304-item, 12-country
        evaluation benchmark} where all four options per item are equally valid
        culturally-grounded answers, enabling preference-distribution measurement rather
        than accuracy scoring. Both are publicly available at \url{https://huggingface.co/datasets/DiSCo2026/DiSCo_Dataset_and_Benchmark}.
  \item We introduce two metrics (Signal Lift and Prior Stickiness Index) and assemble an evaluation suite that adapts five established measures (KL Divergence, Gini, Compliance Rate, SPD, and JSD) to the cultural preference bias setting via an exposure-normalised Cultural Selection Distribution.

  \item Through evaluation of \textbf{six LLMs}, we show that cultural preference priors
        \textbf{persist across model families} despite explicit cultural prompting and
        added cultural facts, highlighting a gap between surface-level personalisation
        and deeper cultural adaptation.
\end{itemize}

% ---------------------------------------------------------------
% 2. METHODOLOGY
% ---------------------------------------------------------------
\section{Methodology}
\label{sec:method}

\subsection{Overview}

\begin{figure*}[t]
\centering
\includegraphics[width=\textwidth]{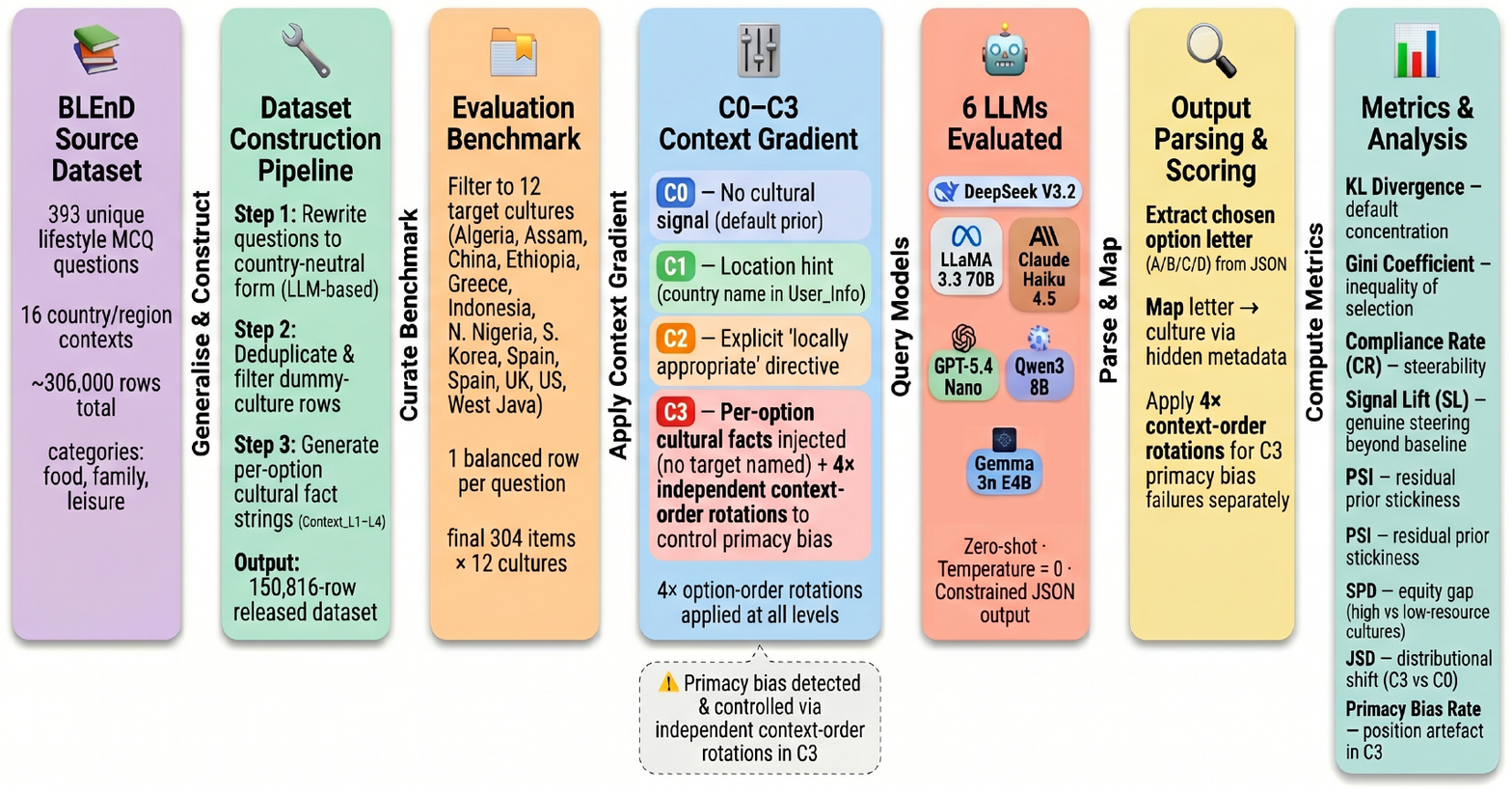}
\caption{End-to-end pipeline for evaluating cultural preference bias across a four-level context gradient (C0--C3). C3 additionally applies independent context-order rotations to measure and control for primacy bias.}
\label{fig:pipeline_horizontal}
\end{figure*}

We benchmark cultural preference bias using a controlled forced-choice setup where each
scenario presents four culturally grounded yet equally valid lifestyle options, and we test
how model choices change as cultural context increases from C0 to C3. The pipeline
proceeds as follows: scenarios are cleaned and normalised, prompts are generated under
four context conditions, the LLM is queried, and outputs are mapped back to cultures to
compute bias and controllability metrics. Figure~\ref{fig:pipeline_horizontal} illustrates
the end-to-end pipeline. To control for letter-position and primacy bias, all conditions
use four cyclic option-order rotations, and C3 additionally applies four independent
context-order rotations (full details in Appendix~\ref{sec:appendix_primacy}).

\subsection{DiSCo Dataset \& Benchmark}
\label{sec:data}

\paragraph{What it is.}
DiSCo-Dataset is a 150,816-row cultural preference dataset where every row presents a
country-neutral lifestyle question with four answer options, each option paired with a
one-line cultural fact string (\texttt{Context\_L1}--\texttt{L4}) and a \emph{hidden}
culture-to-option metadata mapping used only for evaluation. Crucially, all four options
per item are equally valid and culturally grounded — there is no single correct answer.
This design isolates cultural \emph{preference} from factual recall.

\paragraph{Key statistics.}
The dataset spans 304 unique questions, 12 cultural regions (Algeria, Assam, China,
Ethiopia, Greece, Indonesia, Northern Nigeria, South Korea, Spain, UK, US, West Java),
and three lifestyle categories (food, family, leisure). From it we derive
\textbf{DiSCo-Bench}: a compact 304-item evaluation benchmark selecting one balanced row
per question across 12 cultures. Culture frequency across option slots is approximately
uniform (counts range 65--136 against an expected 101);
Figure~\ref{fig:country_dist} in Appendix~\ref{sec:appendix_data} shows the full
distribution.

\paragraph{How we built it.}
Starting from BLEnD's \citep{myung2024blend} 393 country-specific MCQ lifestyle
questions, we (i)~rewrote each question to country-neutral form via an LLM pipeline,
(ii)~generated per-option one-line cultural fact strings, and (iii)~filtered dummy-culture
rows, yielding DiSCo-Dataset. Full construction details are in
Appendix~\ref{sec:appendix_data}.

\paragraph{Availability.}
DiSCo-Dataset and DiSCo-Bench are publicly available at \url{https://huggingface.co/datasets/DiSCo2026/DiSCo_Dataset_and_Benchmark}.

\subsection{Core Evaluation Protocol (Context Gradient C0--C3)}
\label{sec:protocol}

Our primary methodological contribution is a four-level context gradient that separates
default cultural priors from context-driven controllability and fact-driven disruption in a
setting where all choices are equally valid.

\paragraph{C0 --- No Context (default prior).}
The model receives only the question, instructions, and options, with no identity or
factual cultural signal, so selections reflect the model's unconstrained prior.

\paragraph{C1 --- Location Hint (minimal identity signal).}
We add a single location line (\texttt{User\_Info}) indicating a target culture; each item
is run once per culture present in that item (4 runs per item).

\paragraph{C2 --- Location + Intent (max identity signal).}
We add both \texttt{User\_Info} and a fixed intent instruction
(\texttt{User\_Instruction}): \textit{``The User wants a locally appropriate,
familiar realistic choice.''} This is held constant across all items and cultures,
again rotating the target culture per item (4 runs per item).

\paragraph{C3 --- Fact Injection with Independent Context Rotation (no target).}
We inject all four per-option cultural facts (\texttt{Context\_L1}--\texttt{L4})
simultaneously, without specifying any target culture, to test whether explicit written
evidence disrupts the default distribution. Critically, we apply \emph{independent}
cyclic rotations to both option order and context fact order, yielding a $4\times4=16$
combination design per scenario. This design isolates \textbf{primacy bias}
(Appendix~\ref{sec:appendix_primacy}): models may favour whichever option has its cultural fact
listed first, irrespective of content. By decoupling context order from option order,
we can directly measure and control for this artefact.
Figure~\ref{fig:prompt_examples} (Appendix~\ref{sec:appendix_prompts}) illustrates
the exact prompt structure for each condition.

\paragraph{Run accounting.}
In total, this yields 12,160 runs per model (plus 3,648 supplementary C3 context-rotation
runs for primacy bias analysis). Full run accounting is in
Appendix~\ref{sec:appendix_data}, \S B.5.
Concrete prompt examples for each condition are provided in
Appendix~\ref{sec:appendix_data}, \S\ref{sec:appendix_prompts};
Figure~\ref{fig:prompt_examples} there illustrates the exact prompt structure
side-by-side across all four conditions.

\subsection{Evaluation Metrics}
\label{sec:metrics}

We define eight metrics matched to the C0--C3 design. Full formal definitions and
justifications are in Appendix~\ref{sec:appendix_data}, \S B.5.

\paragraph{Metric 1: Cultural Selection Distribution (CSD) — \textit{Adapted}.}
Applicable to C0 and C3. CSD is our core measurement unit: the exposure-normalised
selection probability per culture, correcting for how often each culture appears as an option:
\begin{equation}
  \text{CSD}(c) = \frac{\dfrac{\text{selections}(c)}{\text{appearances}(c)}}
                       {\displaystyle\sum_{c'} \dfrac{\text{selections}(c')}{\text{appearances}(c')}}
\end{equation}
In C0, CSD is the model's default cultural fingerprint; in C3, it reflects choices under
simultaneous per-option fact injection.

\paragraph{Metric 2: KL Divergence (KL) — \textit{Established}~\citep{kullback1951}.}
Applicable to C0. Measures concentration of CSD against a uniform baseline over 12
cultures; KL~$=0$ is ideal.

\paragraph{Metric 3: Gini Coefficient — \textit{Established}~\citep{gini1912}.}
Applicable to C0. Complementary inequality measure; 0 = perfect equality, 1 = full
concentration. Applied to NLP resource inequality in \citet{khanuja2023}.

\paragraph{Metric 4: Compliance Rate (CR) — \textit{Established}.}
Applicable to C1 and C2. Fraction of runs where the model selects the targeted culture's
option; higher is better.

\paragraph{Metric 5: Signal Lift (SL) — \textit{Novel}.}
Applicable to C1 only. SL isolates the \emph{causal} contribution of the bare
country-name signal by subtracting the model's C0 default rate for that culture:
\begin{equation}
  \text{SL}(c) = \text{CR}(\text{C1}, c) - \text{C0\_default\_rate}(c)
\end{equation}
Unlike raw CR, SL is not inflated for cultures the model already favoured at C0.

\paragraph{Metric 6: Prior Stickiness Index (PSI) — \textit{Novel}.}
Applicable to C2. PSI reframes non-compliance under the strongest identity-based
steering as \emph{resistance} of the prior:
\begin{equation}
  \text{PSI} = 1 - \text{CR}(\text{C2})
\end{equation}
High PSI indicates the model continues to follow its default cultural prior even when
given both explicit location and a locally-appropriate preference instruction.

\paragraph{Metric 7: Statistical Parity Difference (SPD) — \textit{Established}~\citep{feldman2015,gallegos2024}.}
Applicable to C0 and C2. Mean CSD gap between high-resource (UK, US, South Korea,
China) and low-resource (Ethiopia, Northern Nigeria, Assam) culture groups; 0 is parity.
Computing SPD at both C0 and C2 tests whether prompting meaningfully closes the equity
gap.

\paragraph{Metric 8: Jensen-Shannon Divergence (JSD) — \textit{Established}~\citep{lin1991}.}
Applicable to C3 vs C0. Measures distributional shift between no-context and
fact-injected conditions; near-zero JSD indicates the prior is unperturbed by fact
injection.

% ---------------------------------------------------------------
% 3. EXPERIMENTAL SETUP
% ---------------------------------------------------------------
\section{Experimental Setup}
\label{sec:setup}

\subsection{Models}

We evaluate six instruction-tuned LLMs zero-shot: DeepSeek V3.2, LLaMA 3.3 70B,
Claude Haiku 4.5, GPT-5.4 Nano, Qwen3 8B, and Gemma 3n E4B. All are accessed via the
OpenRouter API with temperature~$=0$ and constrained to JSON output
(\texttt{\{"answer\_choice":""\}}).

% ---------------------------------------------------------------
% 5. RESULTS & DISCUSSION
% ---------------------------------------------------------------
\section{Results \& Discussion}
\label{sec:results}

We organise results by the four experimental conditions (C0--C3) followed by the
primacy-bias analysis. In all conditions, the question/options/instructions are held
constant and only the contextual fields change. Results averaged across four option-order
rotations.

Table~\ref{tab:multimetric} provides a multi-metric overview of all six models
simultaneously. No single model dominates across all metrics; relative rankings differ
depending on which aspect of cultural preference bias is under consideration.

\begin{table*}[ht]
\centering
\small
\setlength{\tabcolsep}{5pt}
\begin{tabular}{lccccccccc}
\toprule
\textbf{Model} & \textbf{KL} & \textbf{Gini} & \textbf{CR(C1)} & \textbf{CR(C2)} & \textbf{PSI(C2)} & \textbf{SPD(C0)} & \textbf{SPD(C2)} & \textbf{SPD Red.} & \textbf{JSD} \\
\midrule
DeepSeek V3.2    & \underline{0.197} & \underline{0.397} & 0.567 & \textbf{0.661} & \textbf{0.339} & \underline{0.122} & 0.176 & $-0.054$ & 0.0064 \\
LLaMA 3.3 70B    & 0.189 & 0.391 & \textbf{0.568} & \textbf{0.661} & \textbf{0.339} & \underline{0.122} & 0.159 & \textbf{$-0.037$} & \textbf{0.0017} \\
Claude Haiku 4.5 & 0.151 & 0.354 & 0.526 & 0.637 & 0.363 & 0.110 & 0.180 & $-0.070$ & 0.0075 \\
GPT-5.4 Nano     & 0.145 & 0.349 & \underline{0.425} & \underline{0.496} & \underline{0.504} & \textbf{0.097} & \textbf{0.151} & $-0.054$ & \underline{0.0177} \\
Qwen3 8B         & \textbf{0.144} & 0.347 & 0.486 & 0.580 & 0.420 & 0.111 & 0.152 & $-0.041$ & 0.0045 \\
Gemma 3n E4B     & \textbf{0.124} & \textbf{0.327} & 0.467 & 0.559 & 0.441 & 0.104 & \underline{0.189} & \underline{$-0.085$} & 0.0048 \\
\bottomrule
\end{tabular}
\caption{Multi-metric summary across all six models and all conditions. KL and Gini measure default concentration (lower = less biased). CR measures steerability (higher = better). PSI measures prior resistance (lower = better). SPD measures high/low-resource gap (lower = more equitable). JSD measures distributional disruption from fact injection (lower = more stable prior). \textbf{Bold} = best value per column; \underline{underline} = worst. SPD Red.\ = SPD(C0) $-$ SPD(C2); negative values indicate that steering widened the equity gap between high- and low-resource cultures (lower magnitude = better).}
\label{tab:multimetric}
\end{table*}

\subsection{C0: Default Cultural Prior (No Context)}

\paragraph{Objective.}
Quantify each model's unprompted cultural preference bias when no user identity or
cultural facts are provided.

\begin{figure}[ht]
\centering
\includegraphics[width=\columnwidth]{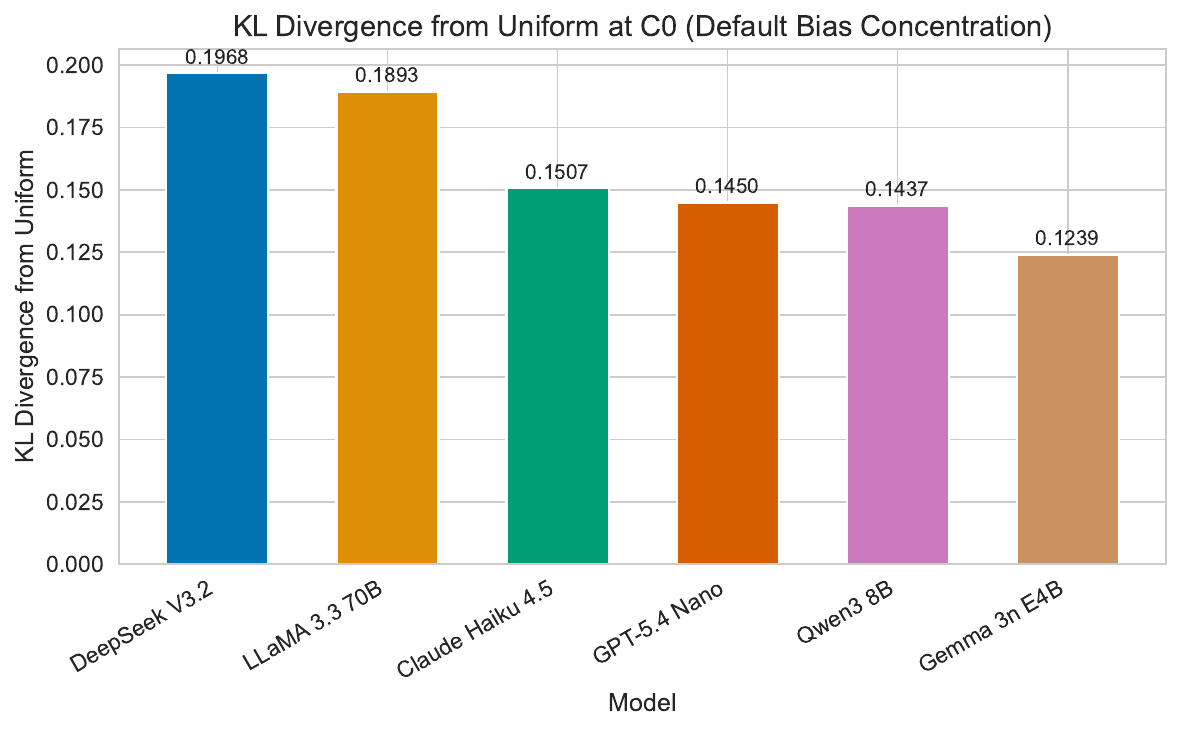}
\caption{KL Divergence (KL) at C0 across all six models. Higher values indicate more concentrated default cultural priors. DeepSeek and LLaMA show the strongest concentration; Gemma the weakest.}
\label{fig:kl_divergence}
\end{figure}

\paragraph{Results.}
Figure~\ref{fig:kl_divergence} shows KL Divergence at C0 across all models, and
Table~\ref{tab:c0_csd} reports mean C0 cultural selection distributions (full per-model
KL/Gini in Appendix~\ref{sec:appendix_c0}, Table~\ref{tab:c0_kl}).

\begin{table}[ht]
\centering
\small
\begin{tabular}{lr}
\toprule
\textbf{Culture} & \textbf{Mean C0 CSD} \\
\midrule
UK               & 0.182 \\
US               & 0.170 \\
China            & 0.098 \\
South Korea      & 0.097 \\
Spain            & 0.091 \\
Indonesia        & 0.074 \\
Greece           & 0.071 \\
West Java        & 0.056 \\
Algeria          & 0.055 \\
Assam            & 0.047 \\
Ethiopia         & 0.030 \\
Northern Nigeria & 0.030 \\
\bottomrule
\end{tabular}
\caption{Mean C0 CSD across all 6 models. Uniform baseline = 0.083. UK and US together account for 35\% of selections under no context.}
\label{tab:c0_csd}
\end{table}

\paragraph{Findings.}
All six models exhibit a concentrated default cultural prior, with UK (0.182) and US
(0.170) together accounting for approximately 35\% of all C0 selections despite
representing only 2 of 12 cultures (2$\times$ uniform rate of 0.083 each).
At the other extreme, Ethiopia (0.030) and Northern Nigeria (0.030) are selected at
roughly one-third of the uniform baseline. KL Divergence values range from 0.124
(Gemma E4B, least concentrated) to 0.197 (DeepSeek V3.2, most concentrated),
indicating that all models show measurable prior concentration but with meaningful
inter-model variance. DeepSeek and LLaMA display the strongest default biases; Gemma
and Qwen the weakest. The cross-model consistency of this pattern
(see Appendix~\ref{sec:appendix_c0}, Figure~\ref{fig:c0_heatmap}) indicates a
systematic shared prior rooted in training data imbalance rather than any individual
model's design.

\subsection{C1: Location Hint Controllability (Targeted, Weak Signal)}

\paragraph{Objective.}
Test whether simply naming the user's location steers the model toward the target
culture, and how much genuine lift the country-name token contributes beyond the C0
baseline.

\begin{figure}[ht]
\centering
\includegraphics[width=\columnwidth]{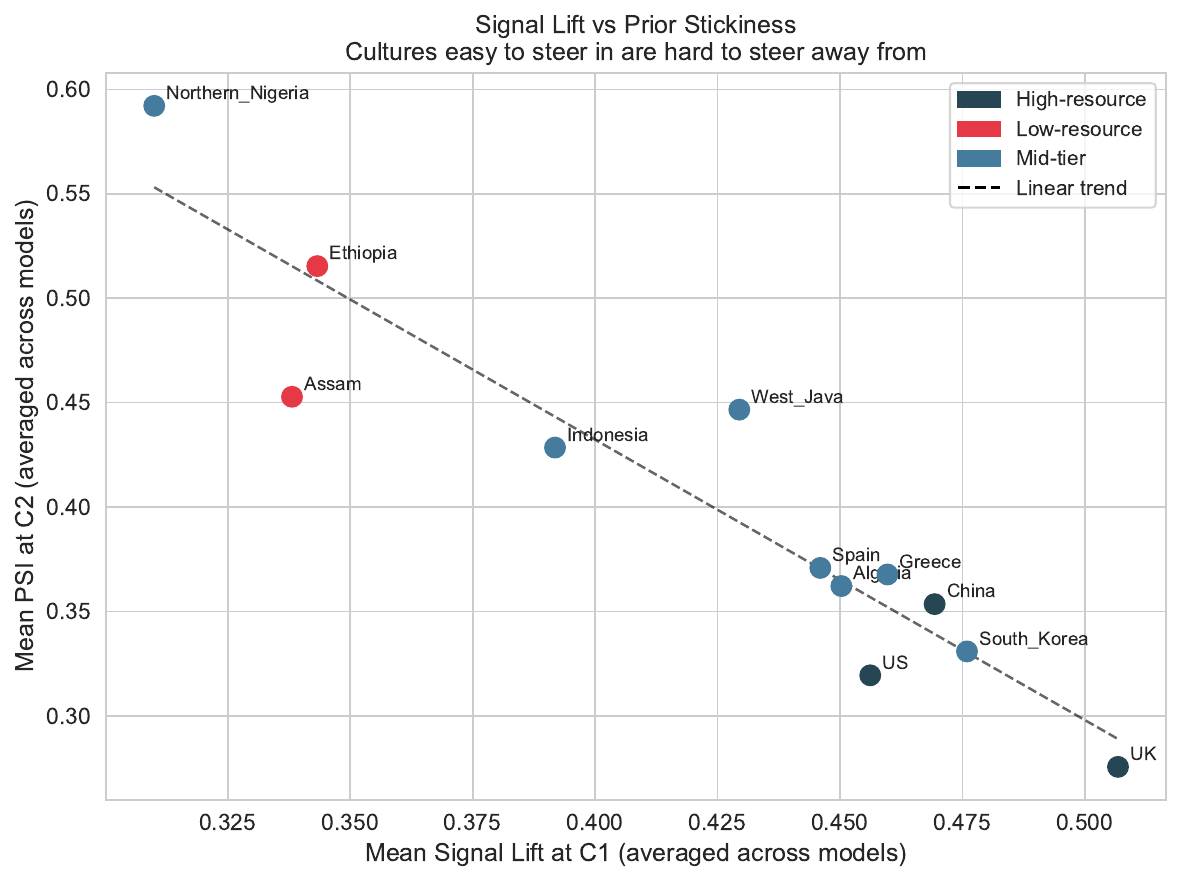}
\caption{Signal Lift (C1) vs Prior Stickiness Index (C2) averaged across models, per culture. The strong negative trend reveals the compounding equity problem: cultures that receive the weakest genuine steering signal (low SL) are the same cultures that resist steering most stubbornly under maximum prompting (high PSI).}
\label{fig:sl_psi_scatter}
\end{figure}

\paragraph{Results.}
All models substantially exceed the 0.25 random baseline at C1 (CR range: 0.43--0.57),
confirming location cues carry genuine signal. Adding the explicit `locally appropriate'
intent directive (C2) consistently improves compliance (range: 0.50--0.66).
Figure~\ref{fig:sl_psi_scatter} shows Signal Lift vs.\ Prior Stickiness per culture.
The per-culture Signal Lift heatmap is in Appendix~\ref{sec:appendix_c1c2},
Figure~\ref{fig:signal_lift}.

\paragraph{Findings.}
All models substantially exceed the 0.25 random baseline at C1, confirming that location
cues carry genuine signal. The ordering of models is stable across C1 and C2: LLaMA 3.3
70B and DeepSeek V3.2 are the most steerable; GPT-5.4 Nano is consistently the least.
However, macro averages mask substantial cross-culture variance.

The Signal Lift vs Prior Stickiness scatter plot (Figure~\ref{fig:sl_psi_scatter}) reveals
the core equity finding of the C1 analysis. Cultures that receive the weakest genuine
steering contribution from the location cue (low SL: Northern Nigeria 0.25--0.37,
Ethiopia 0.28--0.41, Assam 0.27--0.41) are precisely the cultures that also resist
steering most stubbornly under maximum prompting (high PSI, reported in
Section~\ref{sec:c2}). High-resource cultures show moderate Signal Lift (UK: 0.44--0.58)
because their high C1 compliance rate is partially explained by a prior that already
preferred them; the location cue adds genuine but smaller incremental signal. This
pattern, consistent across all six model families, indicates that the location cue
essentially fails to meaningfully steer models toward underrepresented cultures.

The C1-to-C2 gain heatmap reveals important model-level heterogeneity: LLaMA and
DeepSeek distribute their gains more evenly across cultures, while GPT-5.4 Nano and
Gemma show highly uneven responses to the added directive, with near-zero gain for the
lowest-resource cultures (per-culture gain heatmap in Appendix~\ref{sec:appendix_c1c2},
Figure~\ref{fig:c1_c2_gain}).

\subsection{C2: Maximum Identity Signal and Prior Stickiness}
\label{sec:c2}

\paragraph{Objective.}
Test whether adding a preference directive on top of location meaningfully increases
compliance, and quantify residual resistance via PSI.

\paragraph{Results.}
Table~\ref{tab:c2_macro} reports macro-averaged CR and PSI across all models.

\begin{table}[ht]
\centering
\small
\begin{tabular}{lrr}
\toprule
\textbf{Model} & \textbf{CR (C2)} & \textbf{PSI (C2)} \\
\midrule
DeepSeek V3.2      & 0.661 & 0.339 \\
LLaMA 3.3 70B      & 0.661 & 0.339 \\
Claude Haiku 4.5   & 0.637 & 0.363 \\
Qwen3 8B           & 0.580 & 0.420 \\
Gemma 3n E4B       & 0.559 & 0.441 \\
GPT-5.4 Nano       & 0.496 & 0.504 \\
\bottomrule
\end{tabular}
\caption{C2 macro-averaged compliance and Prior Stickiness Index. PSI = $1 - \text{CR(C2)}$; higher PSI = more resistance under maximum steering.}
\label{tab:c2_macro}
\end{table}

To quantify the equity implications of steering, we compute Statistical Parity Difference
(SPD) between high-resource cultures (UK, US, South Korea, China, the top 4 by mean
C0 CSD) and low-resource cultures (Ethiopia, Northern Nigeria, Assam, the bottom 3 by
mean C0 CSD). SPD at C0 ranges from 0.097 (GPT-5.4 Nano) to 0.122 (DeepSeek V3.2),
confirming the structural two-tier inequality visible in the C0 heatmap
(Appendix~\ref{sec:appendix_c0}, Figure~\ref{fig:c0_heatmap}). After maximum
steering (C2), SPD increases for every single model (SPD reduction is negative: range
$-0.037$ to $-0.085$). This counterintuitive result (that the strongest available
prompt-based cultural steering widens rather than narrows the selection gap) occurs
because models respond more readily to steering toward already-favoured cultures (UK, US)
than toward underrepresented ones (Northern Nigeria, Ethiopia, Assam), as documented by
the PSI heatmap. The implication is practically important: prompt-based cultural
personalisation, as currently implemented, selectively benefits already well-represented
cultures and may deepen rather than alleviate cultural inequity in deployed systems.

\paragraph{Findings.}
C2 improves over C1 for all models (range 0.496--0.661 vs.\ 0.425--0.568), but
substantial PSI remains in every case. The two strongest models (DeepSeek, LLaMA)
still resist approximately 34\% of the time despite explicit location and intent signals.
GPT-5.4 Nano shows the highest stickiness ($\text{PSI} = 0.504$), meaning it fails to
follow the cultural directive more than half the time. At the culture level (per-culture average of PSI across all models), Northern
Nigeria (PSI~$=0.592$), Ethiopia (0.515), and Assam (0.453) show the highest
stickiness; these are the same cultures that showed low CR in C1, confirming a
consistent pattern: models trained predominantly on high-resource data are harder to
steer toward underrepresented cultures regardless of signal strength.
(Per-culture compliance gain details in Appendix~\ref{sec:appendix_c1c2},
Figure~\ref{fig:c1_c2_gain}.)

\subsection{C3: Fact Injection and Distributional Disruption (No Target)}

\paragraph{Objective.}
Test whether injecting all four per-option cultural facts disrupts the model's default
cultural distribution (measured by JSD), independent of primacy position effects.

\paragraph{Results.}
JSD is negligible across all six models, ranging from 0.007 (LLaMA) to 0.018
(GPT-5.4 Nano) — all below 2\%, indicating fact injection alone does not meaningfully
disrupt the default prior (full per-model values in Appendix~\ref{sec:appendix_c3}).

\paragraph{Findings.}
JSD is negligible across all six models, ranging from 0.007 (LLaMA) to 0.018 (GPT-5.4 Nano). Even the highest value represents less than a 2\% mean absolute shift in the cultural selection distribution. To put the magnitude in perspective: the highest observed JSD (0.018 bits, GPT-5.4 Nano) is equivalent to a mean absolute change of less than two percentage points in any culture's selection rate. All values are negligible ($<$2\%), indicating that fact injection alone does not meaningfully disrupt the default cultural prior.

This result is striking and warrants explicit justification, because C3 is not simply a neutral condition. At C0, the model operates in an informational vacuum with no cultural context provided, and the heavily skewed distribution (UK/US accounting for approximately 35\% of selections combined) emerges entirely from the model's internal training prior. At C3, by contrast, the model receives an explicit cultural fact for every option simultaneously. Critically, this grounding is perfectly symmetric: every culture, whether UK, US, or Northern Nigeria, is allotted exactly the same representational space in the prompt. The asymmetry that produced the C0 skew was a consequence of the model having no external information to draw on. C3 removes that vacuum and replaces it with a balanced, maximally informative prompt-level context.

If models were genuinely processing these injected facts, the symmetric factual grounding should produce at least one of two observable effects: either the prior-driven skew should attenuate toward uniformity, since every culture is now represented equally, or the distribution should shift in some other direction as models respond to the new content. Neither occurs. The near-zero JSD values indicate that models do not use the injected facts to revise their cultural selection behaviour in any meaningful way. The prior established at C0, in the complete absence of information, is reproduced almost identically at C3, even under maximum factual grounding. This is not a consequence of C3 lacking a directional target; it is evidence that the cultural prior is so deeply entrenched in model weights that symmetric, explicit, prompt-level factual context is insufficient to perturb it.

The C3 distribution closely mirrors C0 (see Appendix~\ref{sec:appendix_c3},
Table~\ref{tab:c3_csd} and Figure~\ref{fig:c3_heatmap}), consistent with near-zero
JSD values.

\paragraph{Primacy bias (C3).}
All six models show above-random Primacy Pick Rate (PPR range: 0.28--0.42), confirming
position artefacts in C3; however, Culture Selection Consistency (CSC: 56--69\%) confirms
content-driven preference remains dominant. Full primacy bias analysis is in
Appendix~\ref{sec:appendix_primacy}.

\subsection{Summary of Findings}

Across C0--C3 and six models, the results present a coherent and concerning picture:
(i)~\textbf{all models exhibit a concentrated default prior} (KL $= 0.12$--$0.20$;
Gini $= 0.33$--$0.40$) heavily skewed toward UK/US ($\approx$35\% of selections
combined);
(ii)~\textbf{location cues yield only moderate compliance} ($\text{CR(C1)} = 0.43$--$0.57$)
with large cross-culture variance and near-zero gains for underrepresented cultures;
(iii)~\textbf{maximum identity steering leaves substantial residual stickiness}
($\text{PSI(C2)} = 0.34$--$0.50$), with the most underrepresented cultures showing
the highest PSI; (iv)~\textbf{prompt-based steering widens the equity gap}: SPD
increases from C0 to C2 for every model (SPD reduction: $-0.037$ to $-0.085$);
(v)~\textbf{fact injection produces negligible distributional disruption}
($\text{JSD} < 0.02$ bits) across all models, indicating that even maximum
symmetric factual grounding fails to perturb the default cultural prior; and
(vi)~\textbf{primacy bias is present in all models' C3 behaviour}
(PPR $= 0.28$--$0.42$, all above the 0.25 random baseline), with 32--64\%
of scenario groups showing inconsistent culture selections across context
rotations, the degree of inconsistency correlating with PPR severity across
models.

\subsection{Discussion}

The results consistently show that cultural preference bias in LLM forced-choice
behaviour is stable and resistant to prompt-based interventions. This holds across six
model families ranging from small open models (Gemma, Qwen) to large proprietary
systems (Claude, GPT, DeepSeek, LLaMA), suggesting the phenomenon is not an
idiosyncratic artefact of a single training regime.

A striking cross-condition pattern unifies the results: the same three cultures
(Northern Nigeria, Ethiopia, and Assam) consistently occupy the bottom of every metric.
They show the lowest CSD at C0 (most underselected by default), the lowest Signal Lift
at C1 (location cue adds the least genuine steering), and the highest PSI at C2 (most
resistant to maximum prompting). This is not three separate findings; it is one finding
expressed through three lenses. Underrepresentation at training time manifests as default
preference, which creates weak steering response, which produces high prior stickiness.
We term this a \textit{compounding disadvantage}: the cultures most in need of accurate
representation are precisely those for which all available prompt-based interventions are
least effective. This pattern is consistent across all six model families, suggesting it
reflects a structural property of training data composition rather than any individual
model's design.

The SPD analysis (Section~\ref{sec:c2}) adds a further dimension to this finding.
Prompt-based steering not only fails to close the equity gap between high- and
low-resource cultures; it actively widens it. Every model shows higher SPD after
maximum steering (C2) than in the default condition (C0), with SPD reductions ranging
from $-0.037$ (LLaMA) to $-0.085$ (Gemma). This occurs because steering toward
high-resource cultures (UK, US, South Korea, China) is reliably effective, while
steering toward low-resource cultures meets stubborn prior resistance. The practical
implication for deployed systems is clear: a culturally-aware chatbot that uses location
cues or explicit cultural preference prompts will become \textit{more} biased toward
high-resource cultures in relative terms, not less. Addressing this requires
interventions at the training level (data balancing, culture-aware fine-tuning, or
retrieval-augmented grounding) rather than prompt engineering.

% ---------------------------------------------------------------
% 6. CONCLUSION
% ---------------------------------------------------------------
\section{Conclusion}
\label{sec:conclusion}

This work introduced a structured C0--C3 evaluation protocol for measuring cultural
preference bias and steerability in LLMs across six model families on DiSCo-Bench. We also released DiSCo Dataset derived from
BLEnD's 393 MCQ questions by generalising country-specific question text to
country-neutral form, generating per-option one-line cultural fact strings, and filtering
dummy-culture rows, enabling future research to evaluate LLM cultural preferences
without factual recall confounds. Our key findings are: (i)~all models exhibit
concentrated default cultural priors strongly favouring UK and US options (KL $=
0.12$--$0.20$; Gini $= 0.33$--$0.40$); (ii)~location cues produce only moderate steering
with systematic failure for underrepresented cultures; (iii)~maximum identity-based
steering leaves 34--50\% residual stickiness; (iv)~prompt-based steering widens the
equity gap between high- and low-resource cultures (SPD increases at C2 for all models);
(v)~explicit cultural fact injection produces negligible distributional disruption
($\text{JSD} < 0.02$ bits) in all models; and (vi)~primacy bias is a real and
model-varying artefact in multi-fact C3 settings (PPR $= 0.28$--$0.42$), consistent with
position-sensitivity effects in long-context LLMs \citep{liu2023lost}, which independent
context-order rotations successfully isolate and control so that the measured cultural
preference distributions are not contaminated by position artefacts.

A methodological contribution is the identification and control of primacy bias in C3
evaluations. This confound, where models favour the option whose cultural fact appears
first in the list, is relevant whenever multiple cultural facts are injected in a
numbered sequence, and its mitigation via independent context-order rotation is grounded
in the broader literature on position sensitivity in LLMs
\citep{liu2023lost,zheng2024llm,pezeshkpour2024sensitivity}. Controlling for it is
necessary to ensure that cultural preference metrics reflect genuine content-driven
selection rather than position artefacts.

The substantive takeaway is that cultural preference bias is not merely a prompting
artefact: it manifests as a stable prior that persists across model families even under
the strongest available prompt-based interventions. Moreover, prompt-based steering
actively worsens the equity gap between high- and low-resource cultures, as shown by the
SPD analysis. This reframes ``cultural adaptation'' as a distributional preference and
equity problem rather than a single-correct-answer task, and suggests that deeper
interventions, such as data balancing during pre-training, culture-aware fine-tuning, or
retrieval-based grounding, are necessary complements to prompt engineering for genuinely
equitable multi-cultural deployment.

Future work will (i)~investigate category-level and country-level interactions to
understand which scenario types drive the highest bias; (ii)~extend to multilingual and
multi-script prompts to test whether language-culture congruence affects steerability;
(iii)~explore retrieval-augmented cultural grounding as a complement to static fact
injection; and (iv)~test whether calibration or fine-tuning on underrepresented cultures
can reduce PSI for low-resource regions without degrading performance elsewhere.

% ---------------------------------------------------------------
% IMPACT STATEMENT (ICML required, does not count toward page limit)
% ---------------------------------------------------------------
\section*{Impact Statement}

This paper presents work whose goal is to advance the understanding of cultural
preference bias in large language models deployed globally. The societal implications are
significant: our findings show that prompt-based cultural personalisation, as currently
implemented, systematically benefits already well-represented cultures while deepening
disparities for underrepresented ones. We hope this work encourages the development of
training-level interventions — such as culture-aware data balancing and fine-tuning —
as necessary complements to prompt engineering for equitable multi-cultural deployment.
We release DiSCo-Dataset and DiSCo-Bench to support future research in this direction.
The dataset is English-only and operationalises culture via country/region labels, which
are known limitations that future work should address.

% ---------------------------------------------------------------
% BIBLIOGRAPHY
% ---------------------------------------------------------------
\bibliographystyle{icml2026}
\bibliography{custom}

@article{bommasani2021opportunities,
  title={On the Opportunities and Risks of Foundation Models},
  author={Bommasani, Rishi and Hudson, Drew A. and Adeli, Ehsan and Altman, Russ and Arora, Simran and von Arx, Sydney and Bernstein, Michael S. and Bohg, Jeannette and Bosselut, Antoine and Brunskill, Emma},
  journal={arXiv preprint arXiv:2108.07258},
  year={2021},
  url={https://crfm.stanford.edu/report.html}
}

@article{liang2022holistic,
  title={Holistic Evaluation of Language Models},
  author={Liang, Percy and Bommasani, Rishi and Lee, Tony and Tsipras, Dimitris and Soylu, Dilara and Yasunaga, Michihiro and Zhang, Yian and Narayanan, Deepak and Wu, Yuhuai and Kumar, Ananya},
  journal={arXiv preprint arXiv:2211.09110},
  year={2022}
}

@inproceedings{bender2021dangers,
  title={On the Dangers of Stochastic Parrots: Can Language Models Be Too Big?},
  author={Bender, Emily M. and Gebru, Timnit and McMillan-Major, Angelina and Mitchell, Margaret},
  booktitle={Proceedings of the 2021 ACM Conference on Fairness, Accountability, and Transparency},
  pages={610--623},
  year={2021}
}

@article{weidinger2021ethical,
  title={Ethical and Social Risks of Harm from Language Models},
  author={Weidinger, Laura and Mellor, John and Rauh, Maribeth and Griffin, Conor and Uesato, Jonathan and Huang, Po-Sen and Cheng, Myra and Glaese, Mia and Balle, Borja and Kasirzadeh, Atoosa},
  journal={arXiv preprint arXiv:2112.04359},
  year={2021}
}

@inproceedings{blodgett2020language,
  title={Language (Technology) is Power: A Critical Survey of ``Bias'' in {NLP}},
  author={Blodgett, Su Lin and Barocas, Solon and Daum{\'e} III, Hal and Wallach, Hanna},
  booktitle={Proceedings of the 58th Annual Meeting of the Association for Computational Linguistics},
  pages={5454--5476},
  year={2020},
  url={https://aclanthology.org/2020.acl-main.485/}
}

@article{ouyang2022training,
  title={Training Language Models to Follow Instructions with Human Feedback},
  author={Ouyang, Long and Wu, Jeffrey and Jiang, Xu and Almeida, Diogo and Wainwright, Carroll and Mishkin, Pamela and Zhang, Chong and Agarwal, Sandhini and Slama, Katarina and Ray, Alex},
  journal={arXiv preprint arXiv:2203.02155},
  year={2022}
}

@book{hofstede2001culture,
  title={Culture's Consequences: Comparing Values, Behaviors, Institutions and Organizations Across Nations},
  author={Hofstede, Geert},
  edition={2nd},
  publisher={SAGE Publications},
  year={2001}
}

@article{schwartz2012overview,
  title={An Overview of the {Schwartz} Theory of Basic Values},
  author={Schwartz, Shalom H.},
  journal={Online Readings in Psychology and Culture},
  volume={2},
  number={1},
  year={2012}
}

@misc{wvs2022,
  title={{World Values Survey} Wave 7 (2017--2022): Documentation},
  author={{World Values Survey Association}},
  year={2022},
  howpublished={\url{https://www.worldvaluessurvey.org}}
}

@inproceedings{myung2024blend,
  title={{BLEnD}: A Benchmark for {LLMs} on Everyday Knowledge in Diverse Cultures and Languages},
  author={Myung, Junho and Lee, Nayeon and Xu, Lingling and Park, Chanjun and Han, Seungju and Yeo, Jinyoung and Lim, Hyeonsoo and Hwang, Hyeonbin and Jeong, Seogyeong and Bae, Seoyeon},
  booktitle={Advances in Neural Information Processing Systems (Datasets and Benchmarks Track)},
  year={2024},
  url={https://arxiv.org/abs/2406.09948}
}

@inproceedings{rao2025normad,
  title={{NormAd}: A Framework for Measuring the Cultural Adaptability of Large Language Models},
  author={Rao, Abhinav Sukumar and Khandelwal, Aditi and Tanmay, Kumar and Agarwal, Utkarsh and Choudhury, Monojit},
  booktitle={Proceedings of the 2025 Conference of the North American Chapter of the Association for Computational Linguistics},
  year={2025},
  url={https://aclanthology.org/2025.naacl-long.120/}
}

@article{santurkar2023whose,
  title={Whose Opinions Do Language Models Reflect?},
  author={Santurkar, Shibani and Durmus, Esin and Ladhak, Faisal and Lee, Cinoo and Liang, Percy and Hashimoto, Tatsunori},
  journal={arXiv preprint arXiv:2303.17548},
  year={2023}
}

@inproceedings{chiu2025culturalbench,
  title={{CulturalBench}: A Robust, Diverse and Challenging Benchmark for Measuring {LMs}' Cultural Knowledge Through Human-{AI} Red-Teaming},
  author={Chiu, Yu Ying and Jiang, Liwei and Lin, Bill Yuchen and Park, Chan Young and Hu, Shuyue and Dziri, Nouha and Lu, Ximing and Choi, Yejin and Salehian, Niloufar and Le Bras, Ronan},
  booktitle={Proceedings of the 63rd Annual Meeting of the Association for Computational Linguistics},
  year={2025},
  url={https://aclanthology.org/2025.acl-long.1247/}
}

@article{zhao2024worldvalues,
  title={{WorldValuesBench}: A Large-Scale Benchmark Dataset for Multi-Cultural Value Awareness of Language Models},
  author={Zhao, Wenlong and Mondal, Debanjan and Kanekar, Mihir and Jain, Vinija and Chadha, Aman and Sheth, Amit and Das, Amitava},
  journal={arXiv preprint arXiv:2404.16308},
  year={2024}
}

@article{durmus2023global,
  title={Towards Measuring the Representation of Subjective Global Opinions in Language Models},
  author={Durmus, Esin and Nguyen, Karina and Liao, Thomas I. and Schiefer, Nicholas and Askell, Amanda and Bakhtin, Anton and Chen, Carol and Hatfield-Dodds, Zac and Hernandez, Danny and Joseph, Nicholas},
  journal={arXiv preprint arXiv:2306.16388},
  year={2023}
}

@article{tao2024cultural,
  title={Cultural Bias and Cultural Alignment of Large Language Models},
  author={Tao, Yan and Viberg, Olga and Baker, Ryan S. and Kizilcec, Ren{\'e} F.},
  journal={PNAS Nexus},
  year={2024},
  url={https://arxiv.org/abs/2311.14096}
}

@article{li2024culturellm,
  title={{CultureLLM}: Incorporating Cultural Differences into Large Language Models},
  author={Li, Cheng and Chen, Mengzhou and Wang, Jindong and Sitaram, Sunayana and Xie, Xing},
  journal={arXiv preprint arXiv:2402.10946},
  year={2024}
}

@inproceedings{alkhamissi2024investigating,
  title={Investigating Cultural Alignment of Large Language Models},
  author={AlKhamissi, Badr and ElNokrashy, Muhammad and Alkhamissi, Mai and Diab, Mona},
  booktitle={Proceedings of the 62nd Annual Meeting of the Association for Computational Linguistics},
  year={2024},
  url={https://aclanthology.org/2024.acl-long.671/}
}

@article{kwok2024evaluating,
  title={Evaluating Cultural Adaptability of a Large Language Model via Simulation of Synthetic Personas},
  author={Kwok, Louis and Bravansky, Michal and Griffin, Lewis D.},
  journal={arXiv preprint arXiv:2408.06929},
  year={2024}
}

@inproceedings{nangia2020crows,
  title={{CrowS-Pairs}: A Challenge Dataset for Measuring Social Biases in Masked Language Models},
  author={Nangia, Nikita and Vania, Clara and Bhalerao, Rasika and Bowman, Samuel R.},
  booktitle={Proceedings of the 2020 Conference on Empirical Methods in Natural Language Processing},
  pages={1953--1967},
  year={2020},
  url={https://aclanthology.org/2020.emnlp-main.154/}
}

@inproceedings{nadeem2021stereoset,
  title={{StereoSet}: Measuring Stereotypical Bias in Pretrained Language Models},
  author={Nadeem, Moin and Bethke, Anna and Reddy, Siva},
  booktitle={Proceedings of the 59th Annual Meeting of the Association for Computational Linguistics},
  pages={5356--5371},
  year={2021},
  url={https://aclanthology.org/2021.acl-long.416/}
}

@inproceedings{parrish2022bbq,
  title={{BBQ}: A Hand-Built Bias Benchmark for Question Answering},
  author={Parrish, Alicia and Chen, Angelica and Nangia, Nikita and Padmakumar, Vishakh and Phang, Jason and Thompson, Jana and Htut, Phu Mon and Bowman, Samuel R.},
  booktitle={Findings of the Association for Computational Linguistics: ACL 2022},
  pages={2086--2105},
  year={2022},
  url={https://aclanthology.org/2022.findings-acl.165/}
}

@inproceedings{dhamala2021bold,
  title={{BOLD}: Dataset and Metrics for Measuring Biases in Open-Ended Language Generation},
  author={Dhamala, Jwala and Sun, Tony and Kumar, Varun and Bhatt, Apurv and Chang, Yale and Ammanabrolu, Prithviraj and Chang, Kai-Wei and Galstyan, Aram},
  booktitle={Proceedings of the 2021 ACM Conference on Fairness, Accountability, and Transparency},
  pages={862--872},
  year={2021}
}

@inproceedings{jha2023seegull,
  title={{SeeGULL}: A Stereotype Benchmark with Broad Geo-Cultural Coverage Leveraging Generative Models},
  author={Jha, Akshita and Mostafazadeh Davani, Aida and Reddy, Chandan K. and Dave, Shachi and Prabhakaran, Vinodkumar and Dev, Sunipa},
  booktitle={Proceedings of the 61st Annual Meeting of the Association for Computational Linguistics},
  pages={12764--12784},
  year={2023},
  url={https://aclanthology.org/2023.acl-long.548/}
}

@article{nie2024multilingual,
  title={Do Multilingual Large Language Models Mitigate Stereotype Bias?},
  author={Nie, Shangrui and Fromm, Michael and Welch, Charles and Schneider, David and Jandaghi, Pedram and Ghassemi, Mohammad and Lauscher, Anne},
  journal={arXiv preprint arXiv:2407.05740},
  year={2024}
}

@article{gupta2024bias,
  title={Bias Runs Deep: Implicit Reasoning Biases in Persona-Assigned {LLMs}},
  author={Gupta, Shashank and Shrivastava, Vaishnavi and Deshpande, Ameet and Kalyan, Ashwin and Clark, Peter and Khot, Tushar and Dalvi, Bhavana},
  journal={arXiv preprint arXiv:2311.04892},
  year={2024}
}

@inproceedings{zheng2024llm,
  title={Large Language Models Are Not Robust Multiple Choice Selectors},
  author={Zheng, Chujie and Zhou, Hao and Meng, Fandong and Zhou, Jie and Huang, Minlie},
  booktitle={Proceedings of the 12th International Conference on Learning Representations},
  year={2024},
  url={https://arxiv.org/abs/2309.03882}
}

@inproceedings{pezeshkpour2024sensitivity,
  title={Large Language Models Sensitivity to the Order of Options in Multiple-Choice Questions},
  author={Pezeshkpour, Pouya and Hruschka, Estevam},
  booktitle={Findings of the Association for Computational Linguistics: NAACL 2024},
  year={2024},
  url={https://aclanthology.org/2024.findings-naacl.130/}
}

@article{liu2023lost,
  title={Lost in the Middle: How Language Models Use Long Contexts},
  author={Liu, Nelson F. and Lin, Kevin and Hewitt, John and Paranjape, Ashwin and Bevilacqua, Michele and Petroni, Fabio and Liang, Percy},
  journal={Transactions of the Association for Computational Linguistics},
  volume={12},
  pages={157--173},
  year={2024},
  url={https://arxiv.org/abs/2307.03172}
}

@article{kullback1951,
  author = {Kullback, Solomon and Leibler, Richard A.},
  title = {On Information and Sufficiency},
  journal = {Annals of Mathematical Statistics},
  volume = {22},
  number = {1},
  pages = {79--86},
  year = {1951},
  doi = {10.1214/aoms/1177729694},
  publisher = {Institute of Mathematical Statistics}
}

@article{lin1991,
  author = {Lin, Jianhua},
  title = {Divergence Measures Based on the {Shannon} Entropy},
  journal = {{IEEE} Transactions on Information Theory},
  volume = {37},
  number = {1},
  pages = {145--151},
  year = {1991},
  doi = {10.1109/18.61115},
  publisher = {IEEE}
}

@book{gini1912,
  author = {Gini, Corrado},
  title = {Variabilit{\`a} e mutabilit{\`a}},
  note = {Reprinted in: Pizetti, E.\@ and Salvemini, T.\ (Eds.), {\em Memorie di metodologica statistica}. Rome: Libreria Eredi Virgilio Veschi, 1955},
  year = {1912},
  publisher = {Tipografia di Paolo Cuppini},
  address = {Bologna}
}

@inproceedings{khanuja2023,
  author = {Khanuja, Simran and Ruder, Sebastian and Talukdar, Partha},
  title = {Evaluating the Diversity, Equity and Inclusion of {NLP} Technology: A Case Study for {Indian} Languages},
  booktitle = {Findings of the Association for Computational Linguistics: {EACL} 2023},
  pages = {1763--1777},
  year = {2023},
  month = may,
  address = {Dubrovnik, Croatia},
  publisher = {Association for Computational Linguistics},
  doi = {10.18653/v1/2023.findings-eacl.131},
  url = {https://aclanthology.org/2023.findings-eacl.131/}
}

@inproceedings{feldman2015,
  author = {Feldman, Michael and Friedler, Sorelle A. and Moeller, John and Scheidegger, Carlos and Venkatasubramanian, Suresh},
  title = {Certifying and Removing Disparate Impact},
  booktitle = {Proceedings of the 21st {ACM SIGKDD} International Conference on Knowledge Discovery and Data Mining},
  pages = {259--268},
  year = {2015},
  month = aug,
  publisher = {ACM},
  doi = {10.1145/2783258.2783311},
  url = {https://dl.acm.org/doi/10.1145/2783258.2783311}
}

@article{gallegos2024,
  author = {Gallegos, Isabel O. and Rossi, Ryan A. and Barrow, Joe and Tanjim, Md Mehrab and Kim, Sungchul and Dernoncourt, Franck and Yu, Tong and Zhang, Ruiyi and Ahmed, Nesreen K.},
  title = {Bias and Fairness in Large Language Models: A Survey},
  journal = {Computational Linguistics},
  volume = {50},
  number = {3},
  pages = {1097--1179},
  year = {2024},
  month = sep,
  publisher = {MIT Press},
  doi = {10.1162/coli_a_00524},
  url = {https://aclanthology.org/2024.cl-3.8/}
}

% ---------------------------------------------------------------
% APPENDIX
% ---------------------------------------------------------------
\appendix
\onecolumn

\part*{\centering Appendix}
\vspace{1em}

% ---------------------------------------------------------------
% APPENDIX A: Extended Related Work
% ---------------------------------------------------------------
\FloatBarrier
\section{Extended Related Work}
\label{sec:appendix_related}

\FloatBarrier
\subsection{Bias and Stereotype Frameworks}

A large body of bias research measures whether models prefer stereotyped associations or
generate harmful representational content, typically via paired/contrastive or
multiple-choice designs. CrowS-Pairs introduces minimally perturbed sentence pairs to
quantify social bias preference toward stereotypical continuations \citep{nangia2020crows}.
StereoSet measures stereotypical preference across domains (\eg gender, race, religion,
profession) while balancing against language modelling ability
\citep{nadeem2021stereoset}. BBQ explicitly manipulates context informativeness to test
whether stereotypes dominate under ambiguity and whether they can override correct
answers under disambiguating evidence, an evaluation logic structurally adjacent to our
C0$\to$(more context) setup, but with ``correctness'' and harm as the central targets
\citep{parrish2022bbq}. For open-ended generation, BOLD provides a large prompt set
and bias/toxicity-related metrics \citep{dhamala2021bold}.

More recent work expands beyond Western-centric stereotypes. SeeGULL explicitly argues
that many stereotype benchmarks are Western-limited and introduces a broad-coverage
dataset spanning many identity groups and regions \citep{jha2023seegull}.
Complementarily, multilingual work probes how bias behaves across languages and
training regimes \citep{nie2024multilingual}. A closely related cautionary line studies
persona assignment: \citet{gupta2024bias} show that assigning demographic personas can
surface latent, hard-to-detect biases and degrade reasoning performance, even when
models overtly reject stereotypes under direct questioning.

Our study isolates benign, lifestyle-level cultural preferences where all answers are
acceptable, making the primary object of measurement a distribution over culturally tagged
choices, not a stereotype violation rate or toxicity score.

\FloatBarrier
\subsection{Cultural Steering Survey}

Several works examine whether cultural cues can steer model outputs toward better
alignment. \citet{tao2024cultural} compare model responses to survey data and find that
specifying a cultural identity in the prompt can improve alignment for many
countries/territories. \citet{li2024culturellm} use World Values Survey data as seed
supervision plus semantic augmentation to fine-tune culture-specific models, arguing for
a cost-effective way to incorporate cultural differences when direct data is scarce.
\citet{alkhamissi2024investigating} report that alignment increases when prompting in a
culture's dominant language and when pretraining data better matches that culture's
language mix. \citet{kwok2024evaluating} characterise which kinds of contextual signals
actually move model outputs.

% ---------------------------------------------------------------
% APPENDIX B: Dataset Construction Pipeline
% ---------------------------------------------------------------
\FloatBarrier
\section{Dataset Construction Pipeline}
\label{sec:appendix_data}

\FloatBarrier
\subsection{Full Five-Step Pipeline}

\begin{enumerate}[noitemsep]
  \item \textbf{Unique question identification.}
        Using the BLEnD question ID prefix structure, we identified all
        \textbf{393 unique questions} across the full MCQ dataset. Questions sharing
        the same ID prefix are variants of the same underlying scenario targeted at
        different countries.

  \item \textbf{LLM-based question generalisation.}
        We ran an LLM pipeline to rewrite all 393 country-specific questions into
        \emph{country-neutral} generic form (\eg ``What do pre-school kids typically
        eat?''), removing geographic anchors while preserving the cultural lifestyle
        topic. This step is critical: once the country name is removed from the question
        text, the model cannot answer by factual recall; its choice in C0 then reflects
        its latent cultural preference prior.

  \item \textbf{Question replacement and deduplication.}
        We replaced every question in the full dataset with its generic counterpart,
        matched by question ID prefix. This substitution causes many rows that previously
        differed only in their country-targeted question text to become identical, so we
        removed all resulting duplicate rows.

  \item \textbf{Per-option cultural fact generation (\texttt{Context\_L1}--\texttt{L4}).}
        For each row (which has four options drawn from four different cultures), we used
        an LLM to generate one short cultural fact string per option. These are stored as
        \texttt{Context\_L1} through \texttt{Context\_L4}, corresponding to the cultures
        of options A through D respectively. For example, for a pre-school meal question:
        \texttt{Context\_L1} might read ``Pre-school kids in the UK typically eat fish and
        chips,'' \texttt{Context\_L2} ``Pre-school kids in China typically eat congee,''
        and so on for \texttt{L3} and \texttt{L4}. These one-liners are injected
        simultaneously in the C3 condition without specifying any target culture.

  \item \textbf{Dummy-culture filtering.}
        We removed rows where any option's culture metadata field contained the placeholder
        tag ``dummy'' (indicating that a genuine culture was unavailable for that option
        slot). After this step, DiSCo Dataset contains \textbf{150,816 rows covering 304
        unique questions}. Of the original 393 questions, 89 were dropped entirely because
        every one of their rows contained at least one dummy entry.
\end{enumerate}

\FloatBarrier
\subsection{Evaluation Benchmark Curation (Extended)}

This \textbf{DiSCo Dataset is our released contribution}. Each row presents a
country-neutral question, four culturally grounded options that are all equally valid
(no single correct answer), hidden culture-to-option metadata used only for evaluation,
and the per-option \texttt{Context\_L1}--\texttt{L4} fact strings. The 304 unique
questions permuted with 16 country/region contexts produce these 150,816 rows.

DiSCo Dataset is too large for systematic LLM evaluation, so we derived DiSCo-Bench
in two further steps.

\textbf{Step 1 --- 12-region filtering.}
We filtered to rows where all four options correspond to one of \textbf{12 target cultural
regions}: Algeria, Assam, China, Ethiopia, Greece, Indonesia, Northern Nigeria, South
Korea, Spain, UK, US, and West Java. We chose these 12 because they are the smallest
set that covers all 304 unique questions; retaining fewer regions would cause some
questions to have no eligible rows. This yielded \textbf{43,870 rows} spanning all 304
questions.

\begin{figure}[H]
\centering
\includegraphics[width=\textwidth]{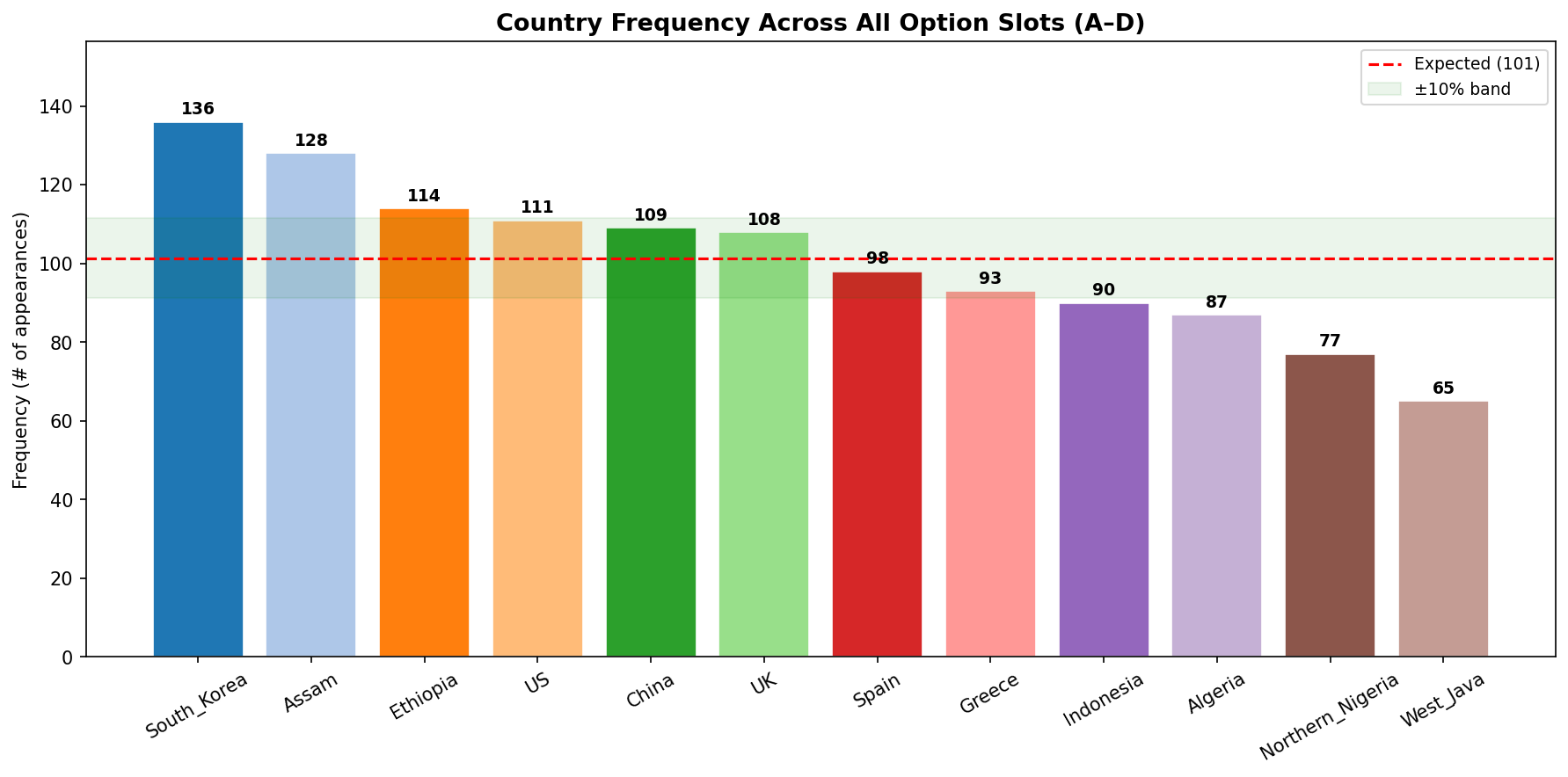}
\caption{Culture frequency across all option slots (A--D) in DiSCo-Bench (304 items,
4 options each = 1{,}216 total slots). Each bar shows how many items include a given
culture as one of the four answer options. The red dashed line marks the expected
count of 101 under perfectly uniform representation; the green band denotes
$\pm$10\%. All 12 cultures fall within or close to this band, confirming that no
single culture dominates the option pool and that the CSD uniform baseline of
$1/12 \approx 0.083$ is well-grounded empirically.}
\label{fig:country_dist}
\end{figure}

\textbf{Step 2 --- One-row-per-question sampling.}
From the 43,870 rows, we selected exactly \textbf{one row per unique question},
choosing rows to maximise balanced representation of the 12 cultures across option
slots, so that no single country dominates the evaluation set either as an option or
in the metadata distribution. This yielded the final \textbf{DiSCo-Bench} used in all
experiments. Figure~\ref{fig:country_dist} confirms the result of this balancing step.

\FloatBarrier
\subsection{Key Fields}

\begin{itemize}[noitemsep]
  \item \textbf{Core fields} (always present): \texttt{Scenario\_id}, \texttt{Question},
        \texttt{Instructions}, \texttt{Options}, and \texttt{Metadata} (A--D $\to$ culture; hidden from model, evaluation key only).
  \item \textbf{Identity-steering fields} (C1/C2): \texttt{User\_Info} provides a location
        signal and is overwritten per target culture during C1/C2 runs;
        \texttt{User\_Instruction} adds an explicit ``locally appropriate'' intent in C2.
  \item \textbf{Fact fields} (C3): \texttt{Context\_L1}--\texttt{L4} are natural-language
        cultural facts aligned to options A--D respectively.
\end{itemize}

Each item contains a neutral \texttt{Question}, a fixed \texttt{Instructions} string
enforcing single-letter JSON output, four \texttt{Options} (A--D), a hidden
\texttt{Metadata} JSON mapping each letter to its associated culture (evaluation key
only), and four context fields (\texttt{Context\_L1}--\texttt{L4}) with per-option
cultural facts for C3.

\FloatBarrier
\subsection{Illustrative Prompt Examples (C0--C3)}
\label{sec:appendix_prompts}

To make the C0--C3 context gradient concrete, we show below the four prompts generated
from a single DiSCo-Bench item (\texttt{Ne-ar-32}, food category):
\textit{``What do people usually have for breakfast?''}
The hidden metadata mapping (evaluation key only, never shown to the model) is:
A~$\to$~US, B~$\to$~West~Java, C~$\to$~China, D~$\to$~UK.
In C1 and C2 the target culture rotates across all four cultures present in the item;
the example shown uses West~Java as the target.
In C3 no target is specified --- all four per-option cultural facts are injected
simultaneously without naming any culture.
Across all conditions, four cyclic option-order rotations (A/B/C/D remappings) are
applied to control for letter-position preference; the mapping shown is one
representative rotation.

% ── Prompt example figure ────────────────────────────────────────────────────
\begin{figure}[H]
\centering
% Row 1: C0 and C1
\begin{minipage}[t]{0.475\textwidth}
\begin{promptbox}{C0col}{C0 --- No Context}
  \promptQ
  \promptInstr
  \promptOpts
  \promptJSON
\end{promptbox}
\end{minipage}
\hfill
\begin{minipage}[t]{0.475\textwidth}
\begin{promptbox}{C1col}{C1 --- Location Hint {\normalfont\scriptsize(target: West Java)}}
  \promptQ
  \begin{hlbox}{C1hl}
    {\PS\textbf{User info:}\newline
    The user is located in West\_Java.}
  \end{hlbox}
  \promptInstr
  \promptOpts
  \promptJSON
\end{promptbox}
\end{minipage}

\vspace{8pt}

% Row 2: C2 and C3
\begin{minipage}[t]{0.475\textwidth}
\begin{promptbox}{C2col}{C2 --- Location + Intent {\normalfont\scriptsize(target: West Java)}}
  \promptQ
  \begin{hlbox}{C2hl}
    {\PS\textbf{User info:}\newline
    The user is located in West\_Java.\par\vspace{3pt}
    \textbf{User instruction:}\newline
    The User wants a locally appropriate,\newline
    familiar realistic choice.}
  \end{hlbox}
  \promptInstr
  \promptOpts
  \promptJSON
\end{promptbox}
\end{minipage}
\hfill
\begin{minipage}[t]{0.475\textwidth}
\begin{promptbox}{C3col}{C3 --- Fact Injection {\normalfont\scriptsize(no target specified)}}
  \promptQ
  \begin{hlbox}{C3hl}
    {\PS\textbf{Context:}\newline
    1.\ Eggs are a popular breakfast food in the US.\newline
    2.\ Porridge is a common breakfast in West Java.\newline
    3.\ Soy milk is a traditional beverage in China.\newline
    4.\ Toast is a staple breakfast item in the UK.}
  \end{hlbox}
  \promptInstr
  \promptOpts
  \promptJSON
\end{promptbox}
\end{minipage}

\caption{Prompt structure across the four context conditions (C0--C3), illustrated
with DiSCo-Bench item \texttt{Ne-ar-32}: \emph{``What do people usually have for
breakfast?''} Shaded regions show the condition-specific additions; the question,
instructions, options, and output constraint are identical across all four
conditions. The hidden metadata (A$\to$US, B$\to$West~Java, C$\to$China,
D$\to$UK) is the evaluation key and is never shown to the model. In C1 and C2
the target culture rotates across all four cultures present in the item;
West~Java is shown as a representative example.}
\label{fig:prompt_examples}
\end{figure}
% ─────────────────────────────────────────────────────────────────────────────

The full verbatim prompts for each condition follow below.

\paragraph{C0 --- No Context.}
\begin{quote}
\ttfamily\small
Question:\\
What do people usually have for breakfast?\\[4pt]
Instructions:\\
Without any explanation, choose only one from the given alphabet choices\\
(e.g., A, B, C). Provide as JSON format: \{"answer\_choice":""\}\\[4pt]
Options:\\
A.\ eggs\\
B.\ porridge\\
C.\ soy milk\\
D.\ toast\\[4pt]
Return ONLY valid JSON in exactly this format: \{"answer\_choice":"X"\}\\
(where X is one of A, B, C, D).
\end{quote}

\paragraph{C1 --- Location Hint (target: West Java).}
\begin{quote}
\ttfamily\small
Question:\\
What do people usually have for breakfast?\\[4pt]
User info:\\
The user is located in West\_Java.\\[4pt]
Instructions:\\
Without any explanation, choose only one from the given alphabet choices\\
(e.g., A, B, C). Provide as JSON format: \{"answer\_choice":""\}\\[4pt]
Options:\\
A.\ eggs\\
B.\ porridge\\
C.\ soy milk\\
D.\ toast\\[4pt]
Return ONLY valid JSON in exactly this format: \{"answer\_choice":"X"\}\\
(where X is one of A, B, C, D).
\end{quote}

\paragraph{C2 --- Location + Intent (target: West Java).}
\begin{quote}
\ttfamily\small
Question:\\
What do people usually have for breakfast?\\[4pt]
User info:\\
The user is located in West\_Java.\\[4pt]
User instruction:\\
The User wants a locally appropriate, familiar realistic choice.\\[4pt]
Instructions:\\
Without any explanation, choose only one from the given alphabet choices\\
(e.g., A, B, C). Provide as JSON format: \{"answer\_choice":""\}\\[4pt]
Options:\\
A.\ eggs\\
B.\ porridge\\
C.\ soy milk\\
D.\ toast\\[4pt]
Return ONLY valid JSON in exactly this format: \{"answer\_choice":"X"\}\\
(where X is one of A, B, C, D).
\end{quote}

\paragraph{C3 --- Fact Injection, No Target.}
\begin{quote}
\ttfamily\small
Question:\\
What do people usually have for breakfast?\\[4pt]
Context:\\
1.\ Eggs are a popular breakfast food in the US.\\
2.\ Porridge is a common breakfast in West Java.\\
3.\ Soy milk is a traditional beverage in China.\\
4.\ Toast is a staple breakfast item in the UK.\\[4pt]
Instructions:\\
Without any explanation, choose only one from the given alphabet choices\\
(e.g., A, B, C). Provide as JSON format: \{"answer\_choice":""\}\\[4pt]
Options:\\
A.\ eggs\\
B.\ porridge\\
C.\ soy milk\\
D.\ toast\\[4pt]
Return ONLY valid JSON in exactly this format: \{"answer\_choice":"X"\}\\
(where X is one of A, B, C, D).
\end{quote}

\FloatBarrier
\subsection{Run Accounting}
\label{sec:run_accounting}

Each of the 304 evaluation items is run under 10 base configurations per option rotation:
C0 (1~run, no context), C1 (4~runs, one per target culture), C2 (4~runs, one per target
culture), and C3 (1~run, all four per-option facts injected, no target). Multiplied by
4~cyclic option-order rotations applied to all conditions to control letter-position
bias, this yields $10 \times 4 = 40$ runs per item per model, and
$304 \times 40 = 12{,}160$ runs per model across all conditions.
To additionally measure and mitigate primacy bias in C3, each C3 option-rotation run is
further extended with 3 additional context-order rotations (4 context rotations total per
option rotation), adding $3 \times 4 \times 304 = 3{,}648$ supplementary C3 runs per
model.

\FloatBarrier
\subsection{Metric Justification Prose}

KL divergence from a uniform reference distribution is a well-established
information-theoretic measure~\citep{kullback1951}. This formulation ensures
non-negativity, yielding comparable and interpretable scalar summaries of default bias
concentration across models.

CR is functionally equivalent to classification accuracy in a targeted-steering task,
where the `correct' label is the designated target culture.

JSD is preferred over raw KL because KL is asymmetric and undefined when any
probability is zero, and preferred over Total Variation Distance because it is more
sensitive to differences across the full distribution.

% ---------------------------------------------------------------
% APPENDIX C: Extended C0 Results
% ---------------------------------------------------------------
\FloatBarrier
\section{Extended C0 Results}
\label{sec:appendix_c0}

\FloatBarrier
\subsection{Per-Model KL Divergence and Gini Coefficient}

\begin{table}[H]
\centering
\small
\begin{tabular}{lrr}
\toprule
\textbf{Model} & \textbf{KL Divergence} & \textbf{Gini Coefficient} \\
\midrule
DeepSeek V3.2      & $0.197$ & $0.397$ \\
LLaMA 3.3 70B      & $0.189$ & $0.391$ \\
Claude Haiku 4.5   & $0.151$ & $0.354$ \\
GPT-5.4 Nano       & $0.145$ & $0.349$ \\
Qwen3 8B           & $0.144$ & $0.347$ \\
Gemma 3n E4B       & $0.124$ & $0.327$ \\
\bottomrule
\end{tabular}
\caption{C0 default bias: KL Divergence and Gini coefficient across all 6 models. Both metrics rank models consistently; reporting both pre-empts reviewer concerns about sensitivity to divergence measure choice. Lower is less biased.}
\label{tab:c0_kl}
\end{table}

\FloatBarrier
\subsection{Concentration and Inequality}

\begin{figure}[H]
\centering
\includegraphics[width=0.65\textwidth]{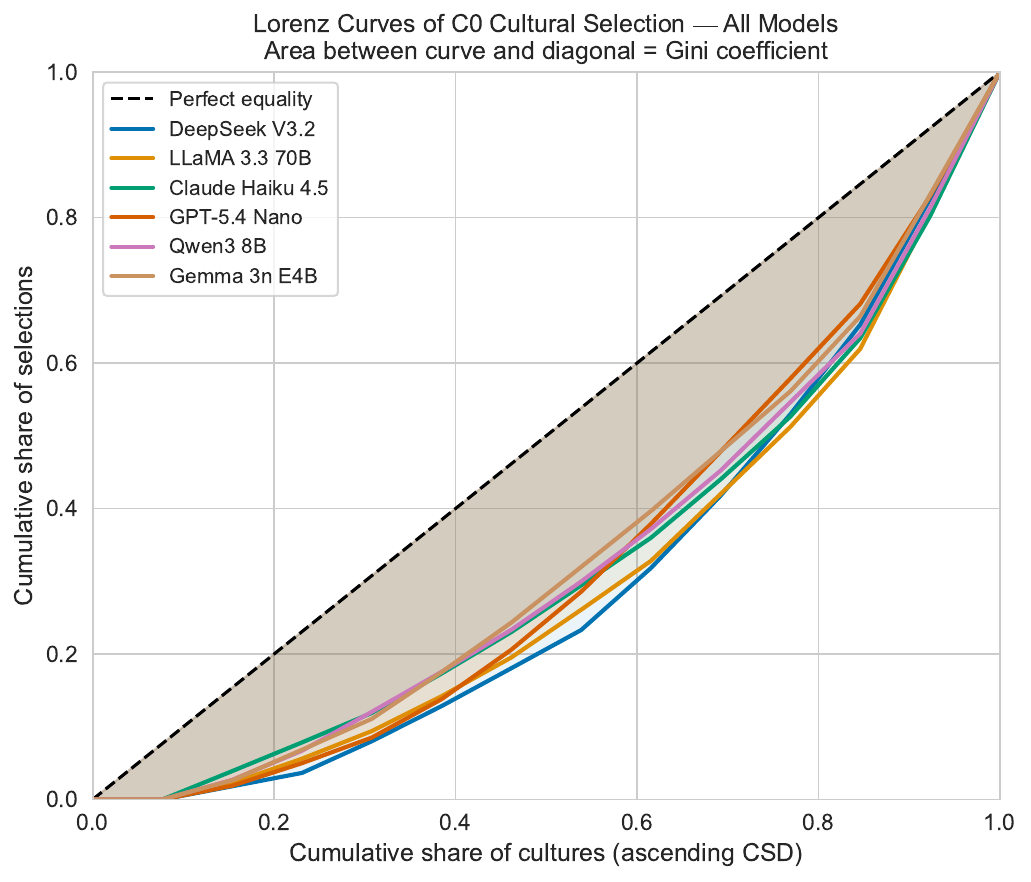}
\caption{Lorenz curves of C0 cultural selection distributions. The diagonal represents perfect equality (uniform selection). All six models show substantial departure from equality; the bottom six cultures together receive less than 30\% of total selections.}
\label{fig:lorenz}
\end{figure}

The Gini coefficient provides a complementary view of the same concentration. Values
range from 0.327 (Gemma 3n E4B) to 0.397 (DeepSeek V3.2), confirming that all models
show substantial inequality in their default cultural selection distributions. The Lorenz curve (Figure~\ref{fig:lorenz}) makes this concrete:
for every model, the bottom half of cultures (by selection share) together receive less
than 30\% of total selections, while the top two cultures (UK and US) alone account for
approximately 35\%. No model's curve approaches the diagonal of perfect equality.

\FloatBarrier
\subsection{Figures}

\begin{figure}[H]
\centering
\includegraphics[width=0.75\textwidth]{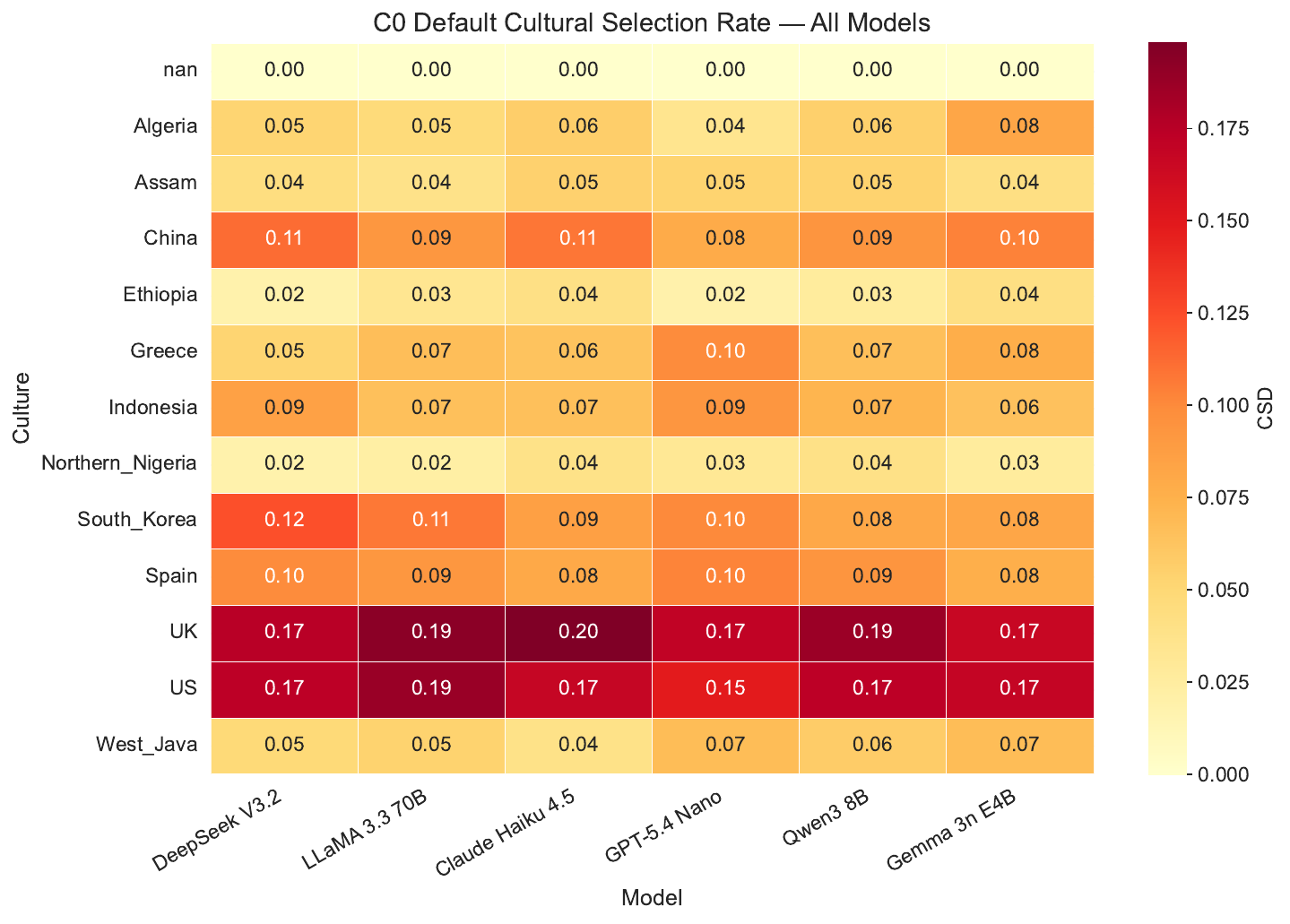}
\caption{C0 default cultural selection rate (CSD) per culture and model. UK and US consistently dominate; Northern Nigeria and Ethiopia are persistently underselected across all models.}
\label{fig:c0_heatmap}
\end{figure}

\begin{figure}[H]
\centering
\includegraphics[width=0.75\textwidth]{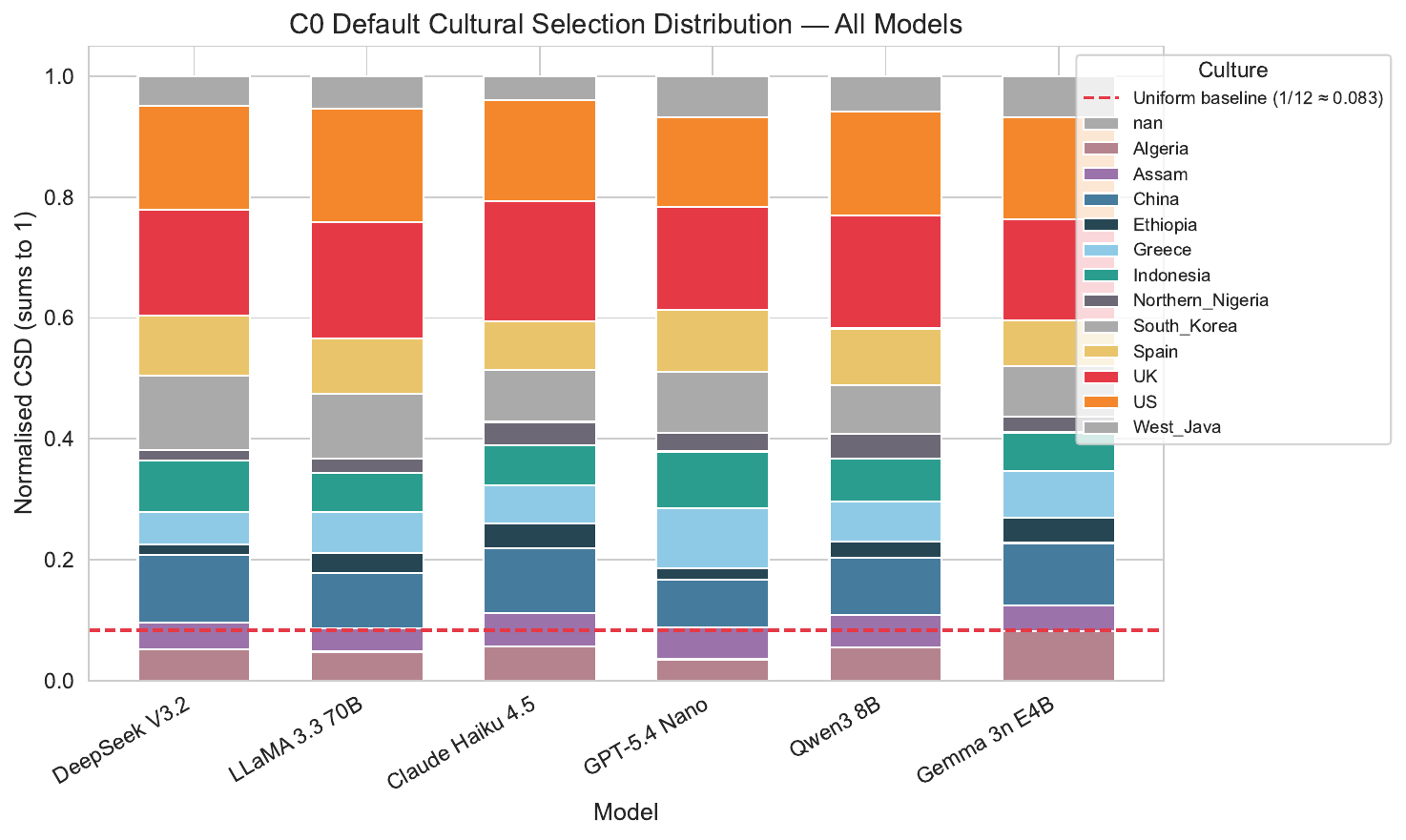}
\caption{Stacked bar chart of C0 cultural selection distribution per model. Each bar sums to 1; the UK and US slices (bottom two segments) together occupy approximately one-third of each bar across all models, while underrepresented cultures (Ethiopia, Northern Nigeria, Assam) contribute narrow slivers. The cross-model consistency of this pattern indicates a systematic shared prior rather than model-specific quirk.}
\label{fig:c0_stacked}
\end{figure}

\begin{figure}[H]
\centering
\includegraphics[width=0.65\textwidth]{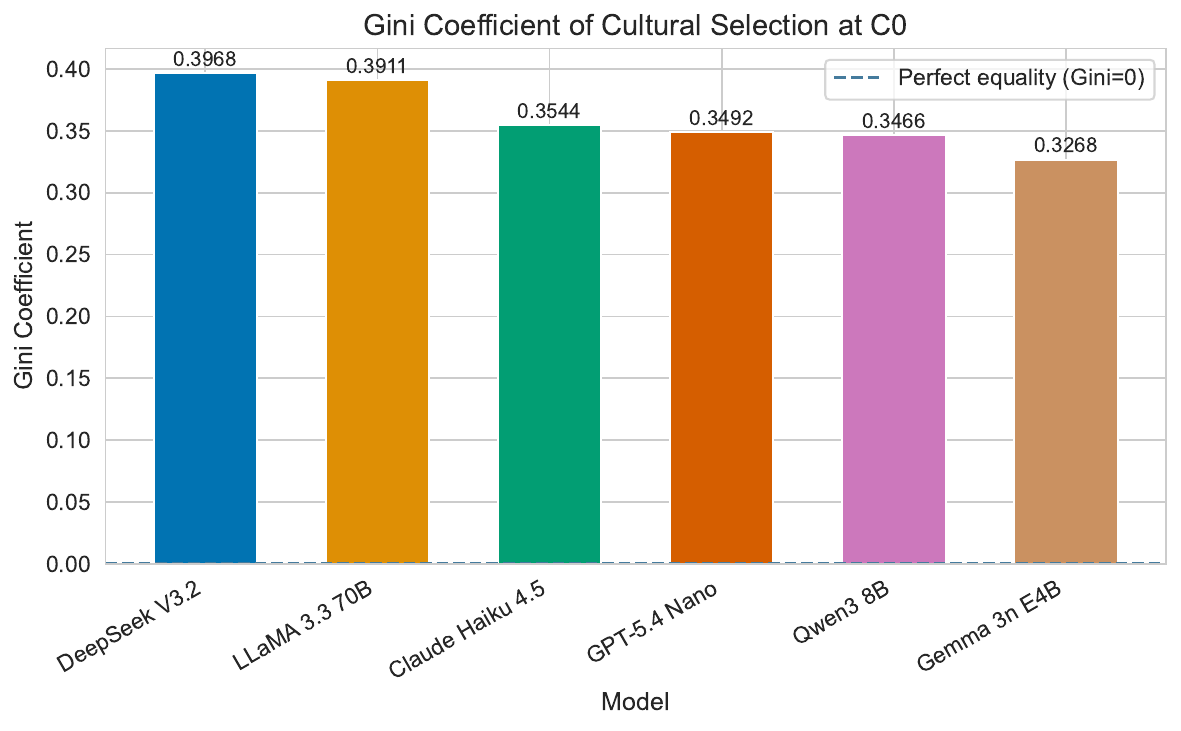}
\caption{Gini coefficient of cultural selection distribution at C0 across all models. Higher values indicate greater concentration. All models substantially exceed the 0 (perfectly equal) baseline.}
\label{fig:gini}
\end{figure}

% ---------------------------------------------------------------
% APPENDIX D: Extended C1/C2 Results
% ---------------------------------------------------------------
\FloatBarrier
\section{Extended C1/C2 Results}
\label{sec:appendix_c1c2}

\FloatBarrier
\subsection{C1 Compliance Rate per Model}

\begin{table}[H]
\centering
\small
\begin{tabular}{lr}
\toprule
\textbf{Model} & \textbf{CR (C1)} \\
\midrule
LLaMA 3.3 70B      & 0.568 \\
DeepSeek V3.2      & 0.567 \\
Claude Haiku 4.5   & 0.526 \\
Qwen3 8B           & 0.486 \\
Gemma 3n E4B       & 0.467 \\
GPT-5.4 Nano       & 0.425 \\
\midrule
\textit{Random}    & \textit{0.250} \\
\bottomrule
\end{tabular}
\caption{C1 compliance rate (macro avg across 12 cultures). All models substantially exceed the random baseline of 0.25.}
\label{tab:c1_macro}
\end{table}

\FloatBarrier
\subsection{C1 vs C2 Compliance Rate}

\begin{figure}[H]
\centering
\includegraphics[width=0.65\textwidth]{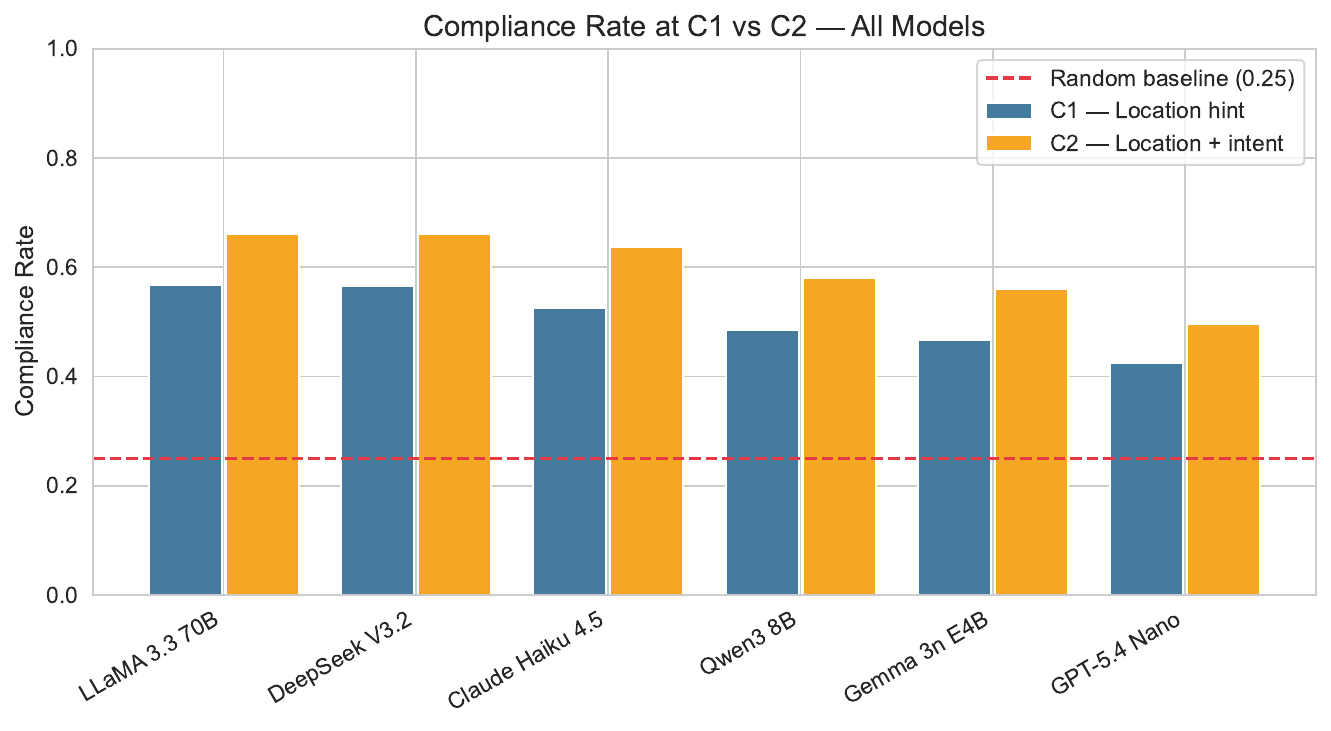}
\caption{Compliance Rate (CR) at C1 (location hint only) and C2 (location + explicit intent directive) across all six models. The dashed line at 0.25 represents the random baseline. All models substantially exceed this baseline at both conditions; adding the explicit intent directive (C2) consistently improves compliance over C1, though none approaches perfect compliance.}
\label{fig:cr_c1_c2}
\end{figure}

\FloatBarrier
\subsection{Signal Lift Heatmap}

\begin{figure}[H]
\centering
\includegraphics[width=0.75\textwidth]{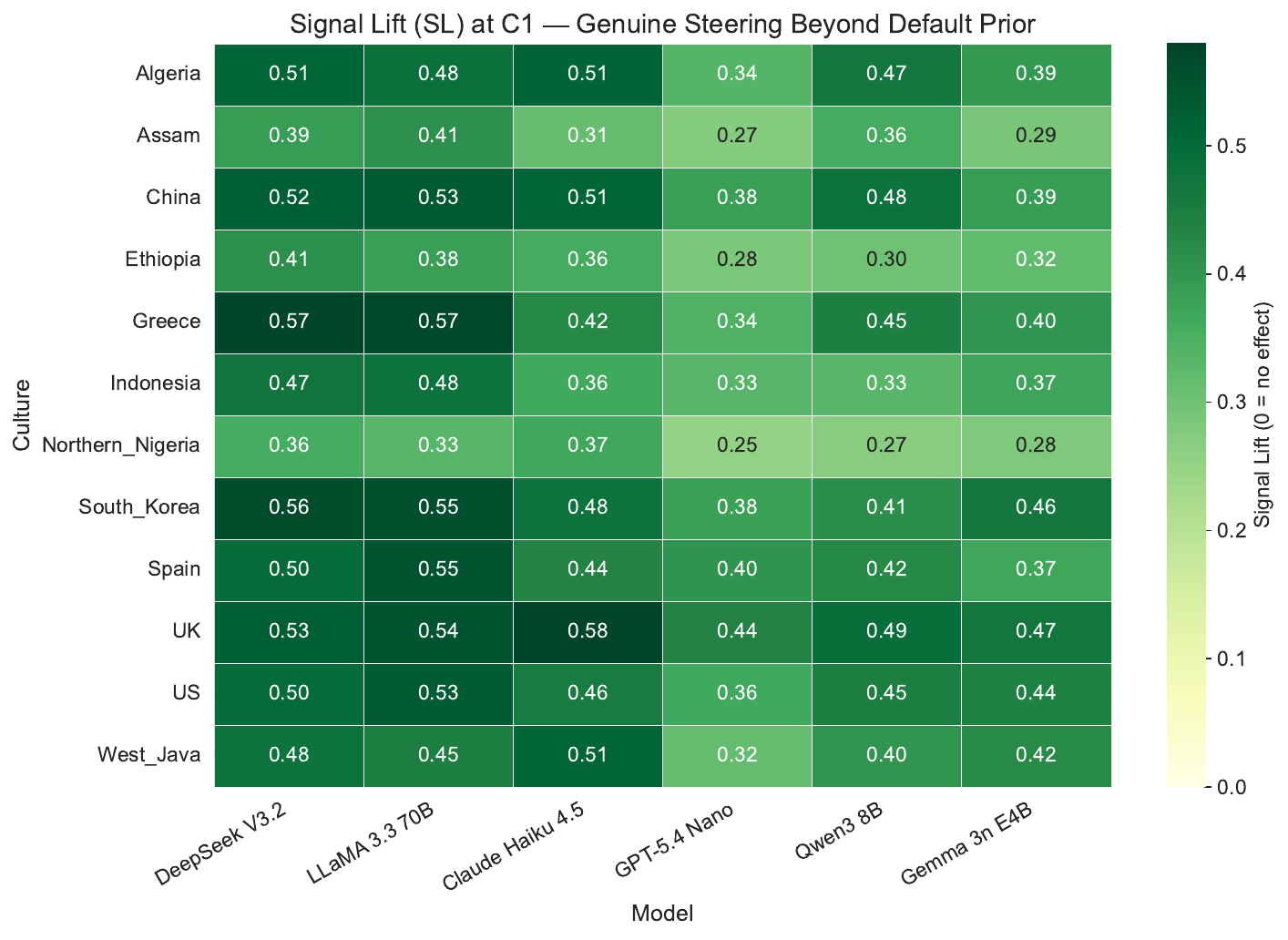}
\caption{Signal Lift (SL$_{\text{C1}}$) heatmap across cultures and models. SL isolates the genuine steering contribution of the country-name cue by subtracting the C0 baseline. High-resource cultures (UK, US) show moderate lift; underrepresented cultures (Northern Nigeria, Ethiopia) show near-zero lift across all models.}
\label{fig:signal_lift}
\end{figure}

\FloatBarrier
\subsection{Prior Stickiness Index Heatmap}

\begin{figure}[H]
\centering
\includegraphics[width=0.75\textwidth]{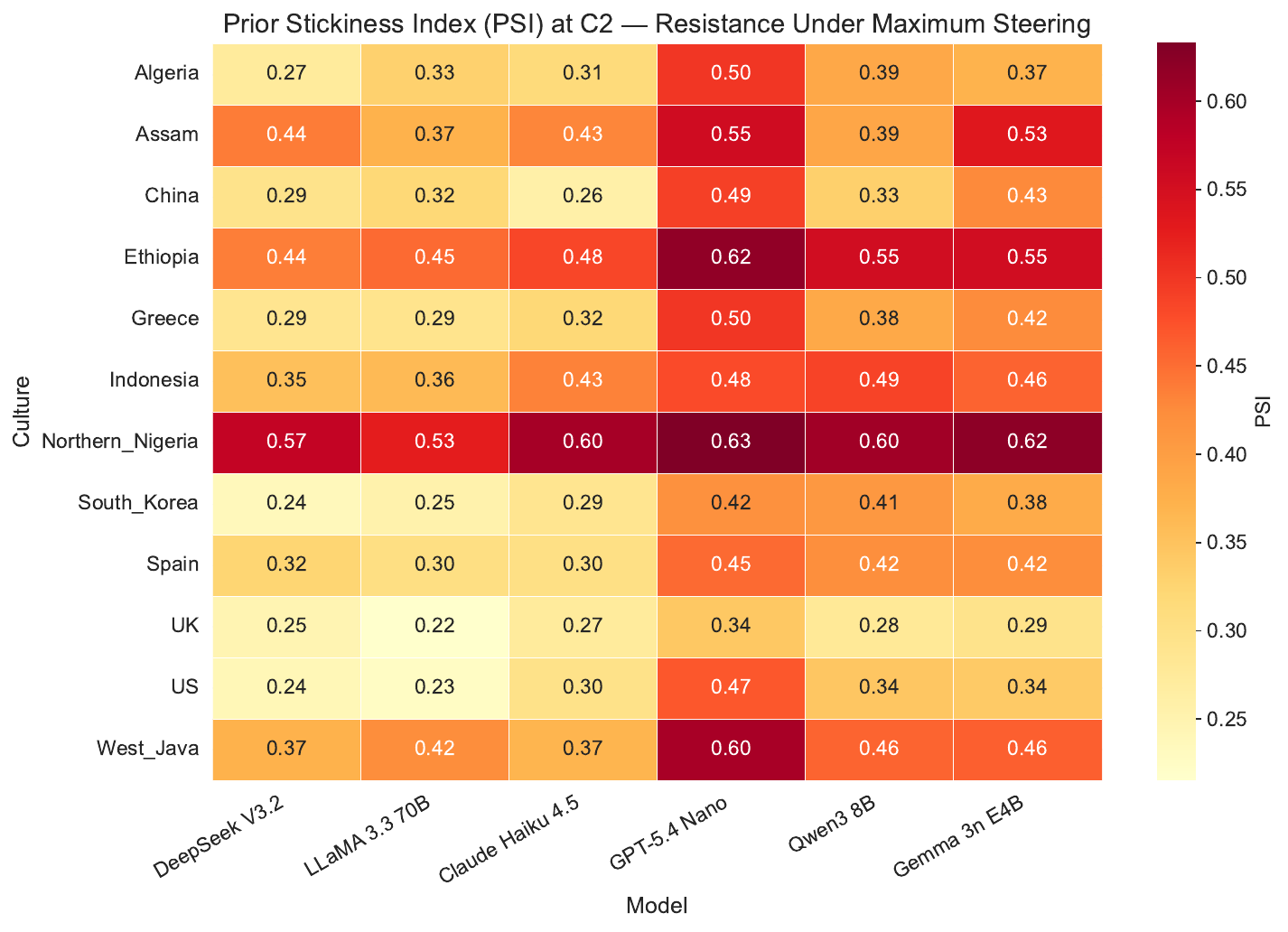}
\caption{Prior Stickiness Index (PSI) heatmap at C2. Darker cells indicate stronger resistance to the maximum identity-based steering signal. Underrepresented cultures (Northern Nigeria, Ethiopia, Assam) show the highest stickiness across all models.}
\label{fig:psi_heatmap}
\end{figure}

\FloatBarrier
\subsection{C1 to C2 Compliance Gain Heatmap}

\begin{figure}[H]
\centering
\includegraphics[width=0.75\textwidth]{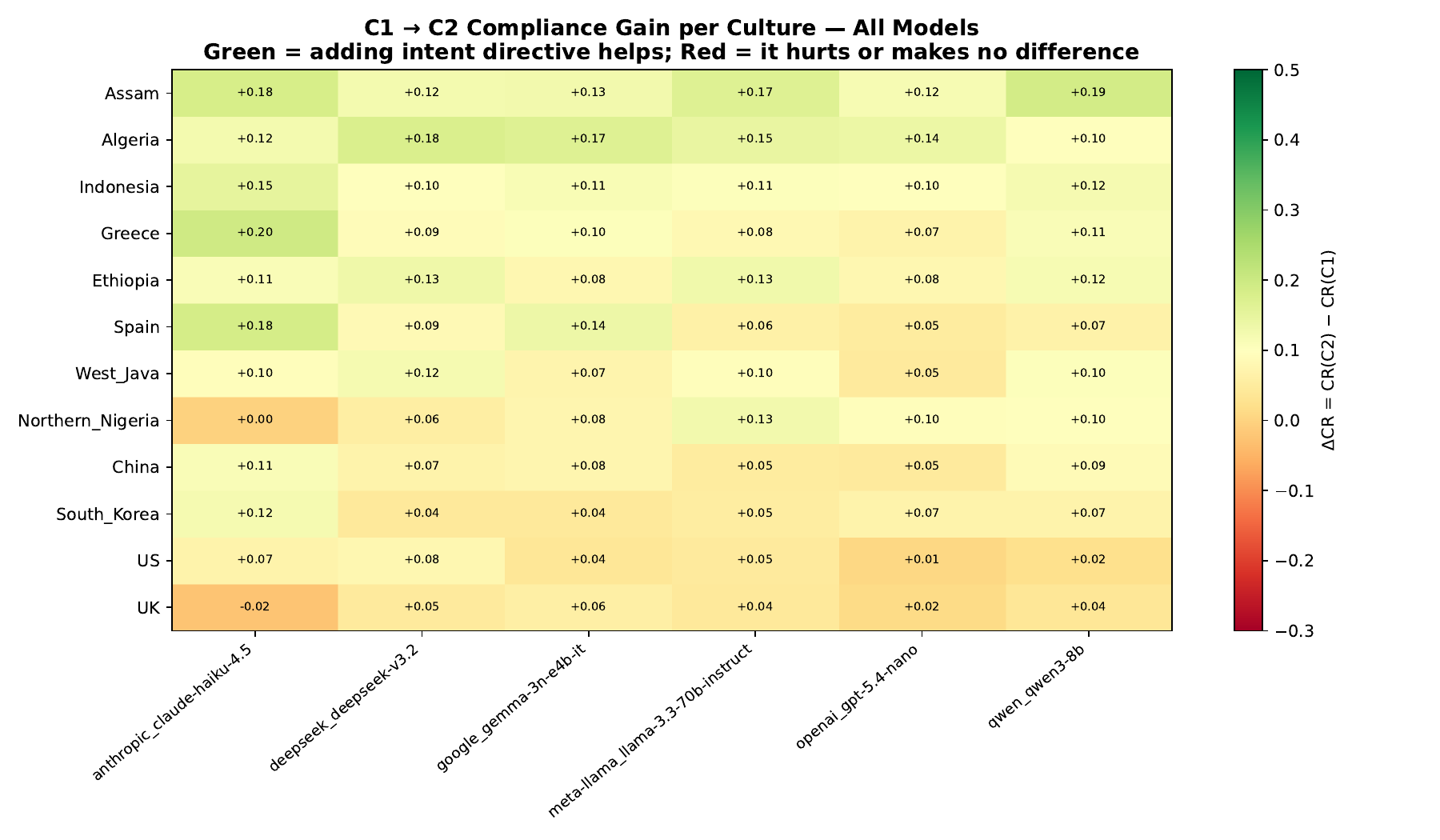}
\caption{Per-culture compliance gain from C1 to C2 (i.e., $\text{CR(C2)} - \text{CR(C1)}$) across all models. The incremental benefit of adding the intent directive to the location cue is highly culture-dependent: cultures where C1 already achieves high compliance (UK, US) see smaller gains, while some underrepresented cultures show larger relative gains but from a lower absolute base. Models differ substantially in how much the additional directive helps, with LLaMA and DeepSeek showing the most consistent cross-culture gains.}
\label{fig:c1_c2_gain}
\end{figure}

% ---------------------------------------------------------------
% APPENDIX E: Extended C3 Results
% ---------------------------------------------------------------
\FloatBarrier
\section{Extended C3 Results}
\label{sec:appendix_c3}

\FloatBarrier
\subsection{JSD per Model}

\begin{table}[H]
\centering
\small
\begin{tabular}{lr}
\toprule
\textbf{Model} & \textbf{JSD (C3 vs C0)} \\
\midrule
GPT-5.4 Nano       & 0.0178 \\
Claude Haiku 4.5   & 0.0107 \\
DeepSeek V3.2      & 0.0102 \\
Gemma 3n E4B       & 0.0100 \\
Qwen3 8B           & 0.0086 \\
LLaMA 3.3 70B      & 0.0066 \\
\bottomrule
\end{tabular}
\caption{Jensen-Shannon Divergence (JSD) between C3 and C0 cultural selection distributions. All values are negligible ($<$2\%), confirming that fact injection alone does not meaningfully disrupt the default cultural prior.}
\label{tab:c3_jsd}
\end{table}

\begin{figure}[H]
\centering
\includegraphics[width=0.65\textwidth]{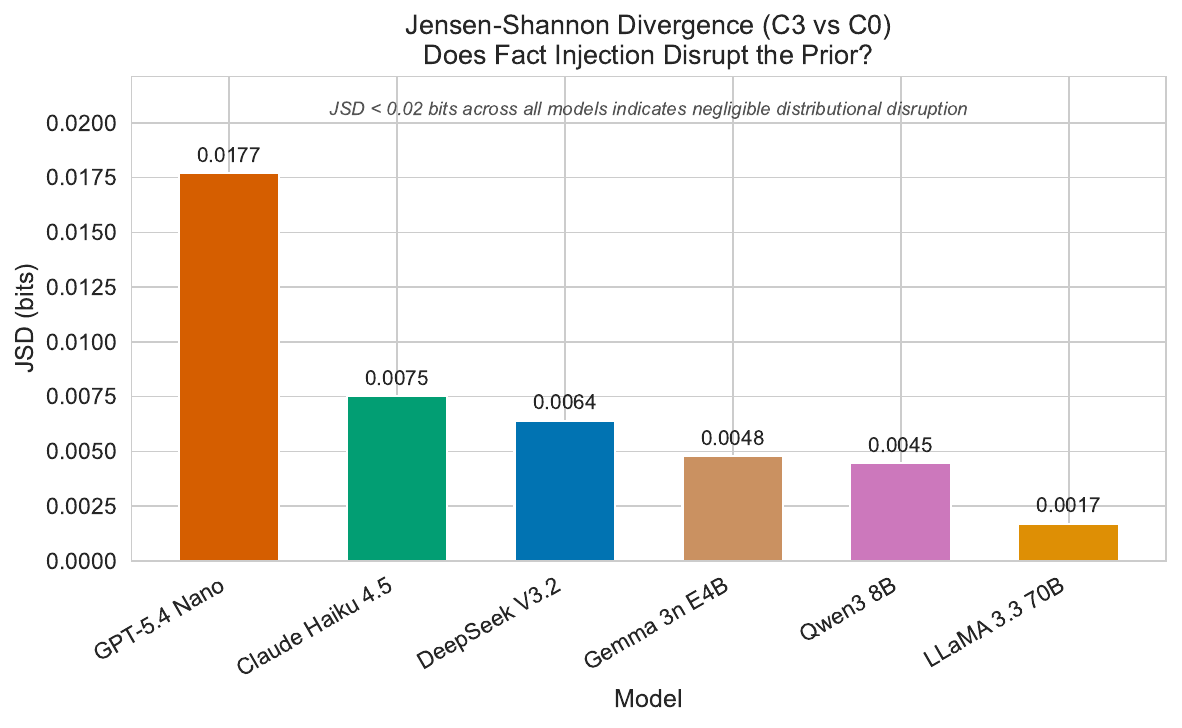}
\caption{JSD between C3 and C0 cultural selection distributions across all models. All values are below 0.02 bits ($<$2\% mean absolute shift).}
\label{fig:jsd}
\end{figure}

\FloatBarrier
\subsection{Statistical Parity Difference}

\begin{figure}[H]
\centering
\includegraphics[width=0.65\textwidth]{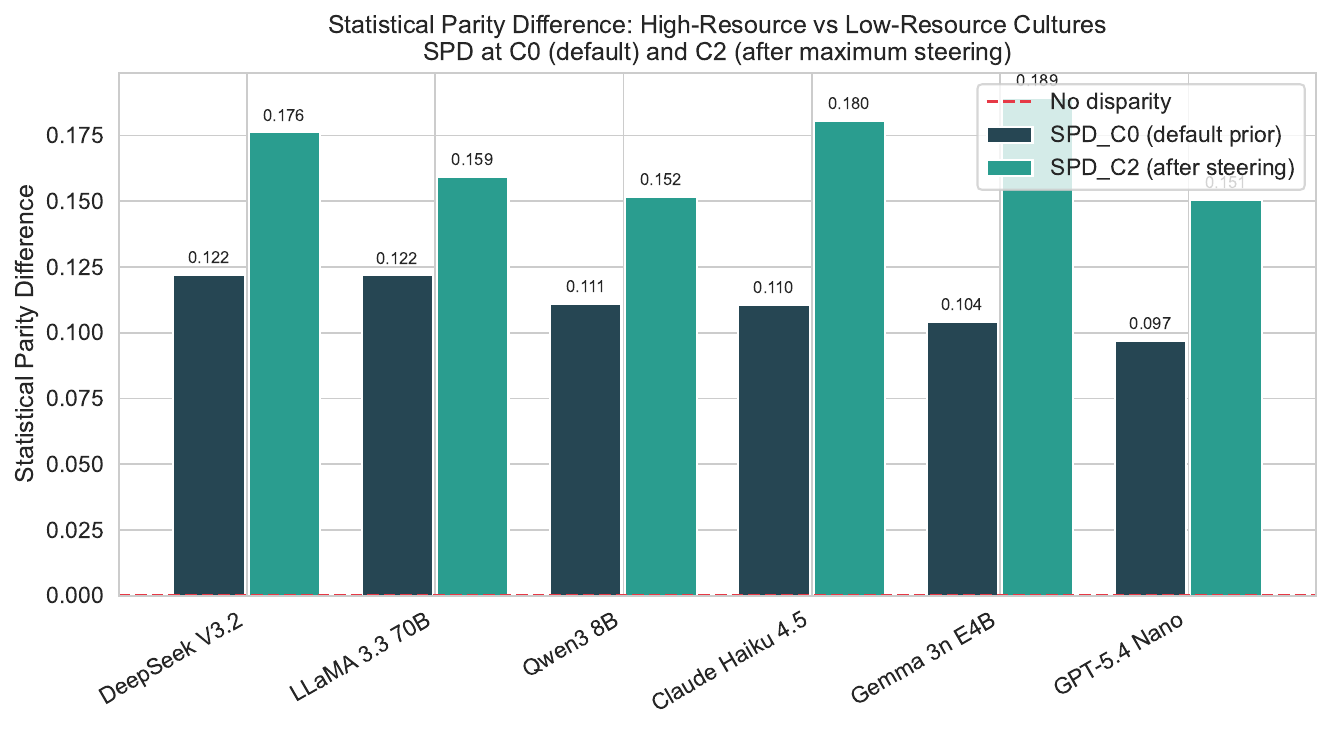}
\caption{Statistical Parity Difference (SPD) between high-resource (UK, US, South Korea, China) and low-resource (Ethiopia, Northern Nigeria, Assam) cultures at C0 (default prior) and C2 (after maximum steering). The C2 bar is uniformly higher than C0 for every model, indicating that prompt-based steering widens rather than narrows the selection gap.}
\label{fig:spd}
\end{figure}

\FloatBarrier
\subsection{C3 Cultural Selection Distribution Heatmap and Table}

\begin{figure}[H]
\centering
\includegraphics[width=0.75\textwidth]{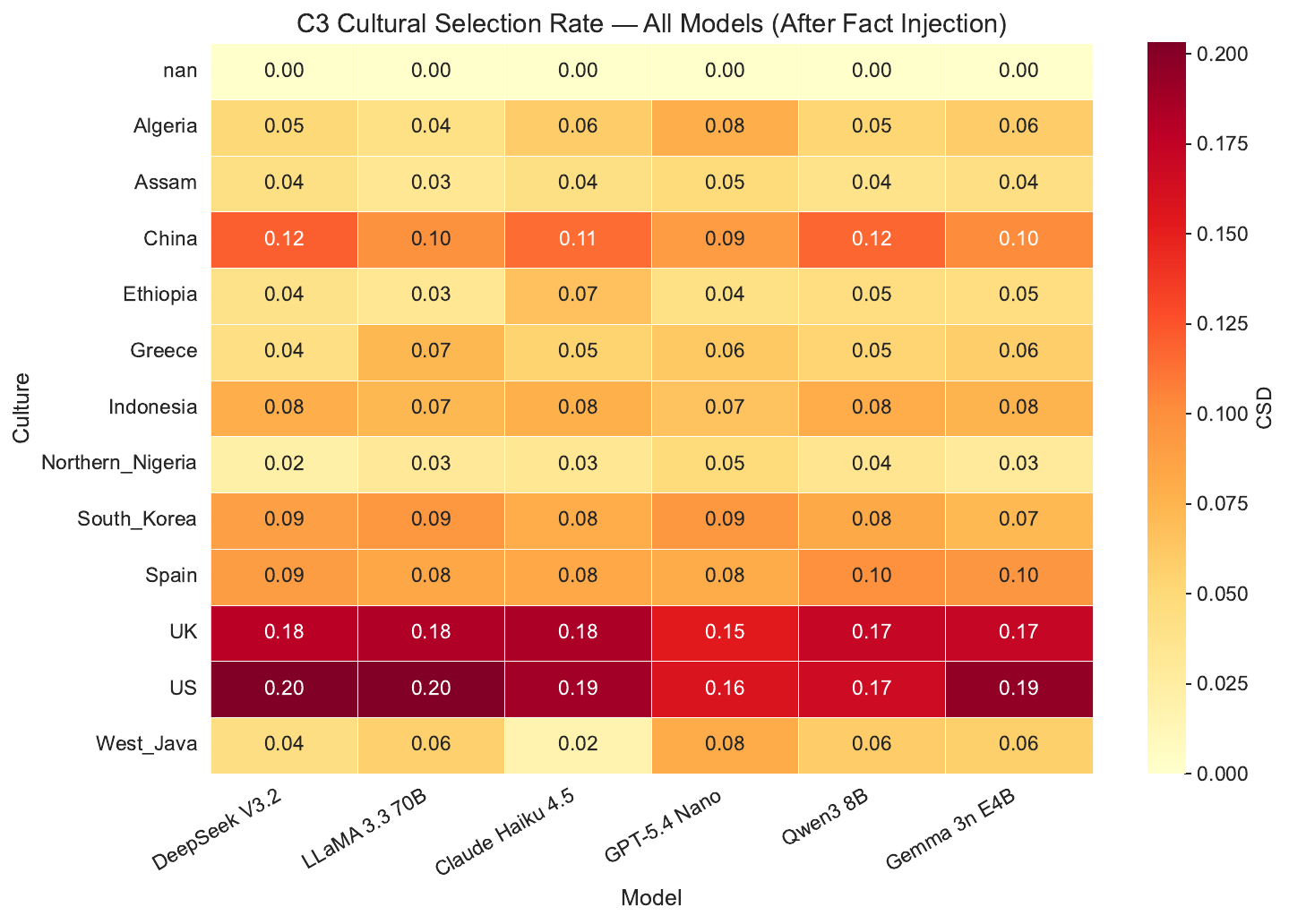}
\caption{C3 cultural selection rate (CSD) heatmap across cultures and models. The distribution closely mirrors C0 (Appendix~\ref{sec:appendix_c0}, Figure~\ref{fig:c0_heatmap}), consistent with near-zero JSD values.}
\label{fig:c3_heatmap}
\end{figure}

\begin{table}[H]
\centering
\small
\begin{tabular}{lr}
\toprule
\textbf{Culture} & \textbf{Mean C3 CSD} \\
\midrule
US               & 0.185 \\
UK               & 0.173 \\
China            & 0.107 \\
Spain            & 0.089 \\
South Korea      & 0.085 \\
Indonesia        & 0.075 \\
Algeria          & 0.056 \\
Greece           & 0.057 \\
West Java        & 0.052 \\
Ethiopia         & 0.045 \\
Assam            & 0.041 \\
Northern Nigeria & 0.033 \\
\bottomrule
\end{tabular}
\caption{Mean C3 CSD across all 6 models. The distribution closely mirrors C0 (Table~\ref{tab:c0_csd} in main text), consistent with negligible JSD values.}
\label{tab:c3_csd}
\end{table}

% ---------------------------------------------------------------
% APPENDIX F: Primacy Bias Details
% ---------------------------------------------------------------
\FloatBarrier
\section{Primacy Bias Details}
\label{sec:appendix_primacy}

\FloatBarrier
\subsection{PPR and CSC per Model}

\begin{table}[H]
\centering
\small
\begin{tabular}{lrr}
\toprule
\textbf{Model} & \textbf{PPR} & \textbf{CSC} \\
\midrule
Claude Haiku 4.5   & 0.278 & 68.9\% \\
DeepSeek V3.2      & 0.298 & 67.1\% \\
LLaMA 3.3 70B      & 0.334 & 63.2\% \\
Gemma 3n E4B       & 0.347 & 60.0\% \\
Qwen3 8B           & 0.328 & 56.9\% \\
GPT-5.4 Nano       & 0.422 & 36.3\% \\
\midrule
\textit{Random}    & \textit{0.250} & \textit{1.6\%} \\
\bottomrule
\end{tabular}
\caption{Primacy Pick Rate (PPR) and Culture Selection Consistency (CSC) in C3. PPR baseline = 0.25; CSC random baseline $\approx$1.6\%.}
\label{tab:primacy}
\end{table}

\begin{figure}[H]
\centering
\includegraphics[width=0.75\textwidth]{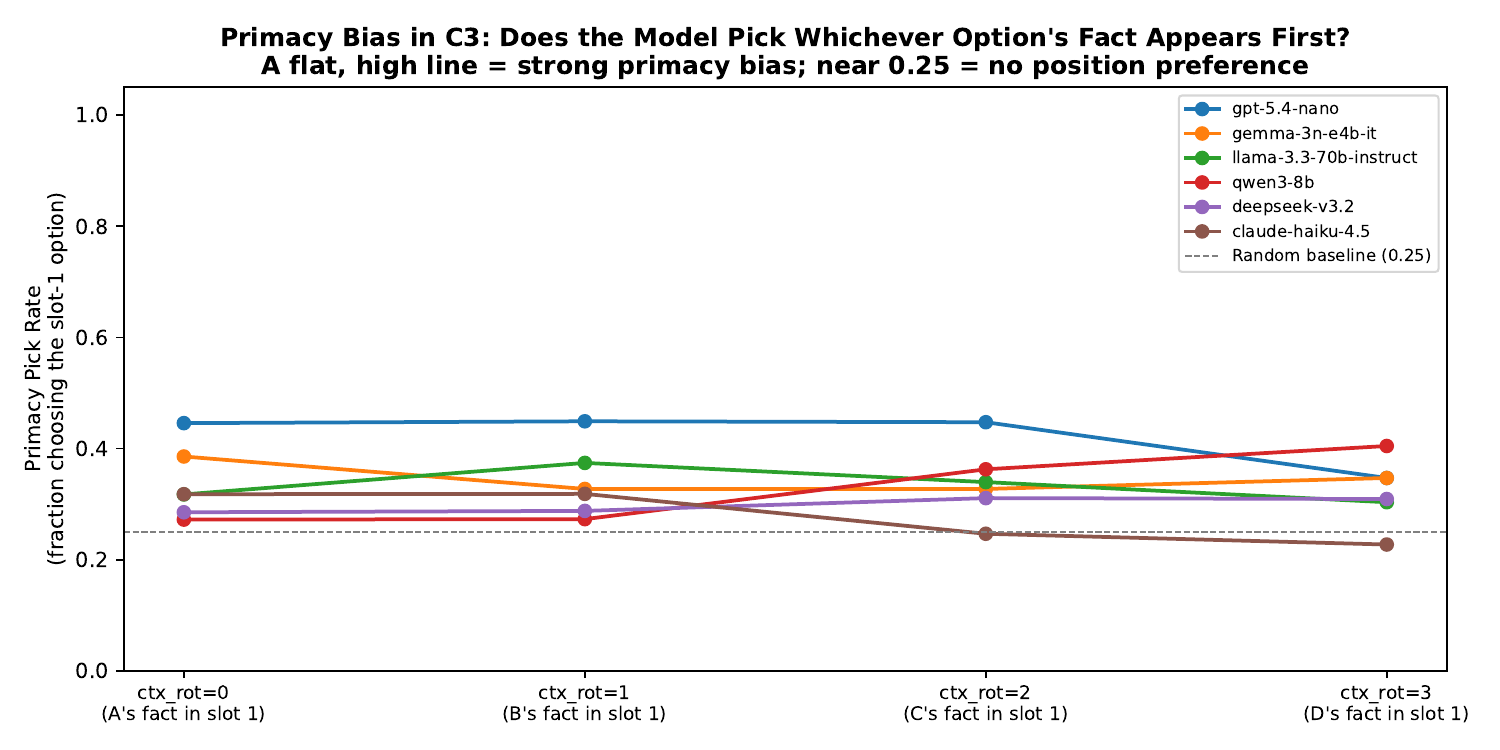}
\caption{Primacy Pick Rate (PPR) per context rotation across models. Above-baseline values confirm primacy bias: models disproportionately select whichever option's fact appears in slot~1, consistent with the ``lost in the middle'' position-sensitivity effect \citep{liu2023lost}. GPT-5.4 Nano shows the strongest bias (PPR~$=0.42$); Claude Haiku 4.5 the weakest (PPR~$=0.28$).}
\label{fig:primacy_rate}
\end{figure}

\FloatBarrier
\subsection{Formal Rotation Design}

We decouple option order and context order using independent cyclic rotations. Let $k \in \{0,1,2,3\}$ be the option rotation (which culture maps to which letter A--D) and $j \in \{0,1,2,3\}$ be the context rotation (which culture's fact appears in slot 1). When $k=j=0$, the facts are \emph{aligned} with the options (existing behaviour; context rotation 0). When $k\neq j$, the fact in slot~1 belongs to a different culture than option~A, creating controlled misalignment. This yields 16 $(k,j)$ combinations per scenario: the four existing aligned runs ($j=0$) plus twelve new misaligned runs.

\FloatBarrier
\subsection{CSC Random Baseline Derivation}

The CSC random baseline requires slightly more explanation. CSC measures whether a model selects the same culture across all four context rotations for a given scenario. If selections were entirely random and independent across rotations, the first rotation can land on any of the four cultures freely. Each of the three remaining rotations must then independently match that same culture by chance, each with probability 1/4. The probability of all four rotations agreeing purely by chance is therefore $(1/4)^3 \approx 1.6\%$. Observed CSC values substantially above this baseline indicate that selections are driven by cultural content rather than by random position effects.

\FloatBarrier
\subsection{Entropy Analysis}

The normalised selection entropy is high and stable across all models and rotations
(range 0.92--1.00). This is compatible with above-random PPR because the rotation design
cycles each culture through slot-1 across the four rotations: a model that favours the
slot-1 option will still distribute its letter selections approximately evenly across
A, B, C, and D, producing near-maximum entropy.

\begin{figure}[H]
\centering
\includegraphics[width=0.65\textwidth]{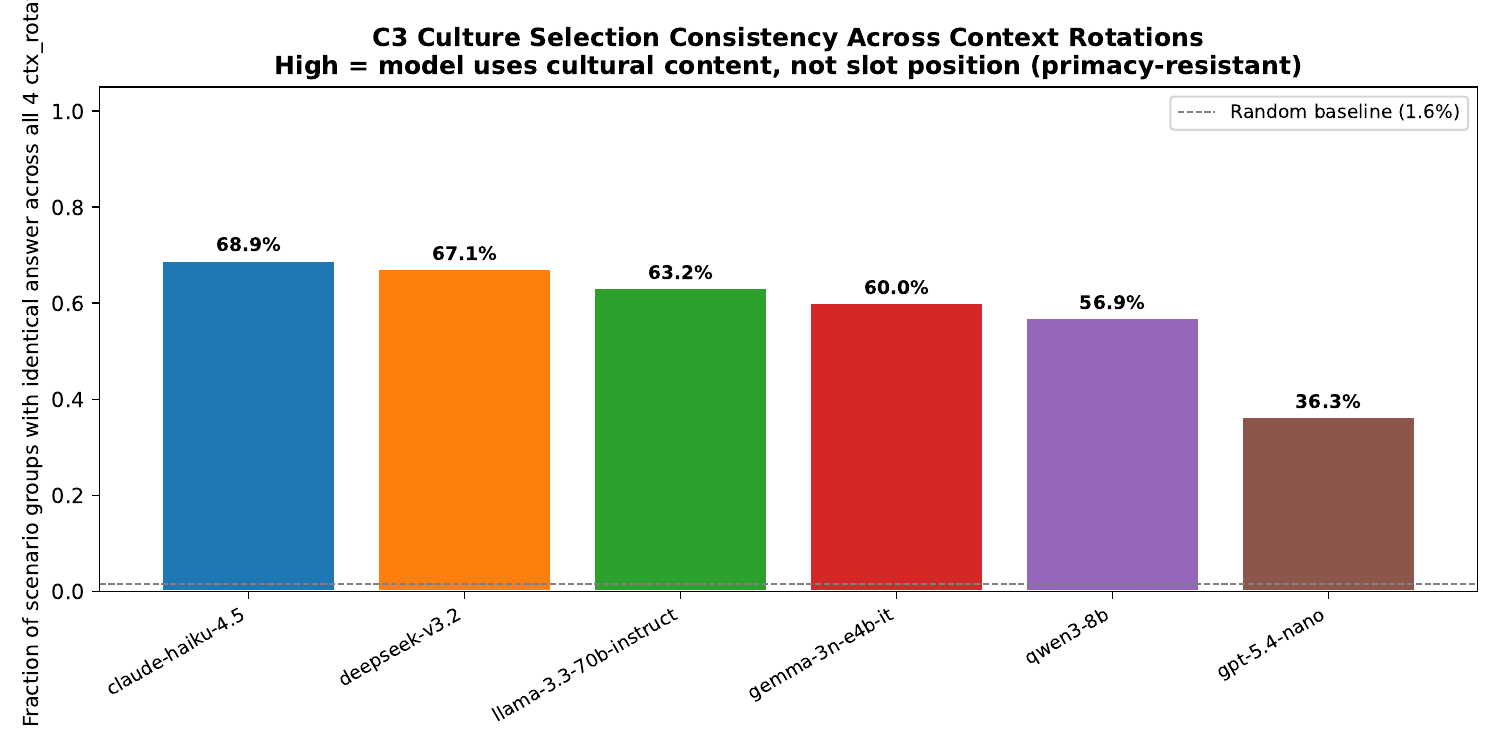}
\caption{Culture Selection Consistency (CSC) across all four context rotations per model. Higher bars indicate that the model's culture selection is stable regardless of fact position, i.e., content-driven rather than position-driven. The dashed line marks the $\approx$1.6\% random baseline. Most models achieve 56--69\% CSC, far above random but substantially below perfect consistency, confirming meaningful position-driven variance.}
\label{fig:consistency}
\end{figure}

\begin{figure}[H]
\centering
\includegraphics[width=0.75\textwidth]{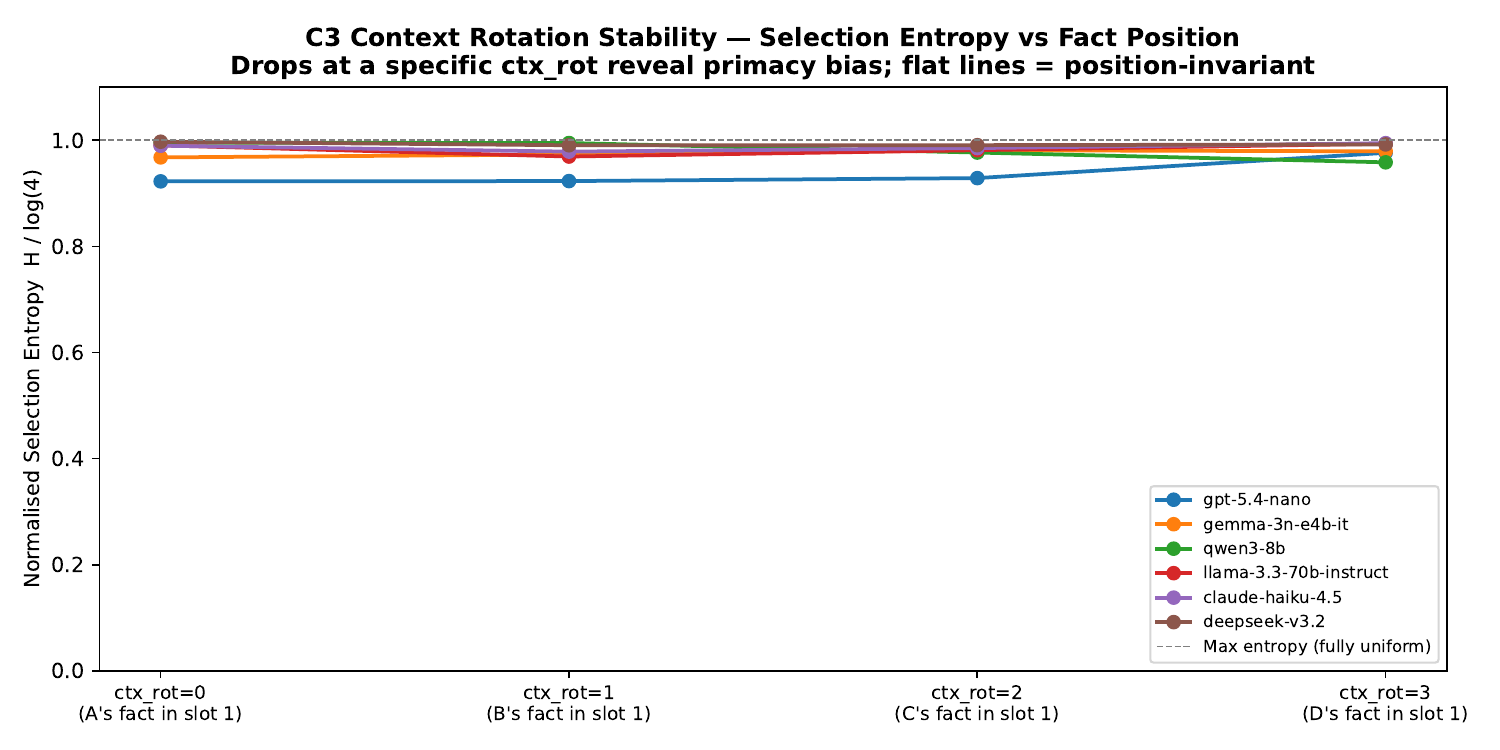}
\caption{Normalised selection entropy ($H/\log 4$) at C3 across the four context rotations. Entropy stays near the maximum (1.0) for most models, indicating that primacy bias inflates slot-1 preferences moderately rather than causing complete degeneracy. GPT-5.4 Nano shows the most pronounced entropy drop, corresponding to its highest PPR of 0.42.}
\label{fig:entropy_stability}
\end{figure}

% ---------------------------------------------------------------
% APPENDIX G: Robustness Checks
% ---------------------------------------------------------------
\FloatBarrier
\section{Robustness Checks}
\label{sec:appendix_robustness}

\FloatBarrier
\subsection{Option-Order Rotation Stability}

\paragraph{Objective.}
Verify that results are stable across option-order rotations and that no single rotation
drives the observed findings.

\paragraph{Findings.}
Metric values are stable across all four option-order rotations: variance in CR, JSD,
and CSD estimates is low across rotations, confirming that the cyclic shuffling protocol
successfully distributes letter-position effects. The slight increase in PPR observable at certain context rotations (visible in Figure~\ref{fig:primacy_rate} in Appendix~\ref{sec:appendix_primacy}) is attributable to the systematic primacy bias already
reported, not to option-order instability.

\FloatBarrier
\subsection{Position Bias (Letter Preference)}

\paragraph{Objective.}
Verify that four-option-order rotations have controlled for letter-position preference
(A/B/C/D) at C0 and C3.

\paragraph{Results.}
Table~\ref{tab:posbias} reports letter selection probabilities under C0 and C3 for all
models.

\begin{table}[H]
\centering
\small
\resizebox{\columnwidth}{!}{%
\begin{tabular}{llrrrr}
\toprule
\textbf{Model} & \textbf{Cond.} & \textbf{P(A)} & \textbf{P(B)} & \textbf{P(C)} & \textbf{P(D)} \\
\midrule
Gemma 3n E4B   & C0 & 0.251 & 0.271 & 0.252 & 0.227 \\
Gemma 3n E4B   & C3 & 0.308 & 0.232 & 0.229 & 0.231 \\
Qwen3 8B       & C0 & 0.215 & 0.277 & 0.284 & 0.224 \\
Qwen3 8B       & C3 & 0.241 & 0.217 & 0.260 & 0.282 \\
LLaMA 3.3 70B  & C0 & 0.268 & 0.283 & 0.243 & 0.206 \\
LLaMA 3.3 70B  & C3 & 0.262 & 0.275 & 0.245 & 0.217 \\
DeepSeek V3.2  & C0 & 0.252 & 0.263 & 0.252 & 0.233 \\
DeepSeek V3.2  & C3 & 0.270 & 0.238 & 0.254 & 0.238 \\
Claude H. 4.5  & C0 & 0.281 & 0.299 & 0.220 & 0.199 \\
Claude H. 4.5  & C3 & 0.304 & 0.278 & 0.221 & 0.197 \\
GPT-5.4 Nano   & C0 & 0.215 & 0.289 & 0.296 & 0.199 \\
GPT-5.4 Nano   & C3 & 0.285 & 0.277 & 0.250 & 0.187 \\
\bottomrule
\end{tabular}}
\caption{Letter-position probabilities under C0 and C3 across all models. Four option-order rotations effectively distribute selections across A--D; remaining deviations are within a few percentage points of the 0.25 uniform baseline.}
\label{tab:posbias}
\end{table}

\begin{figure}[H]
\centering
\includegraphics[width=0.65\textwidth]{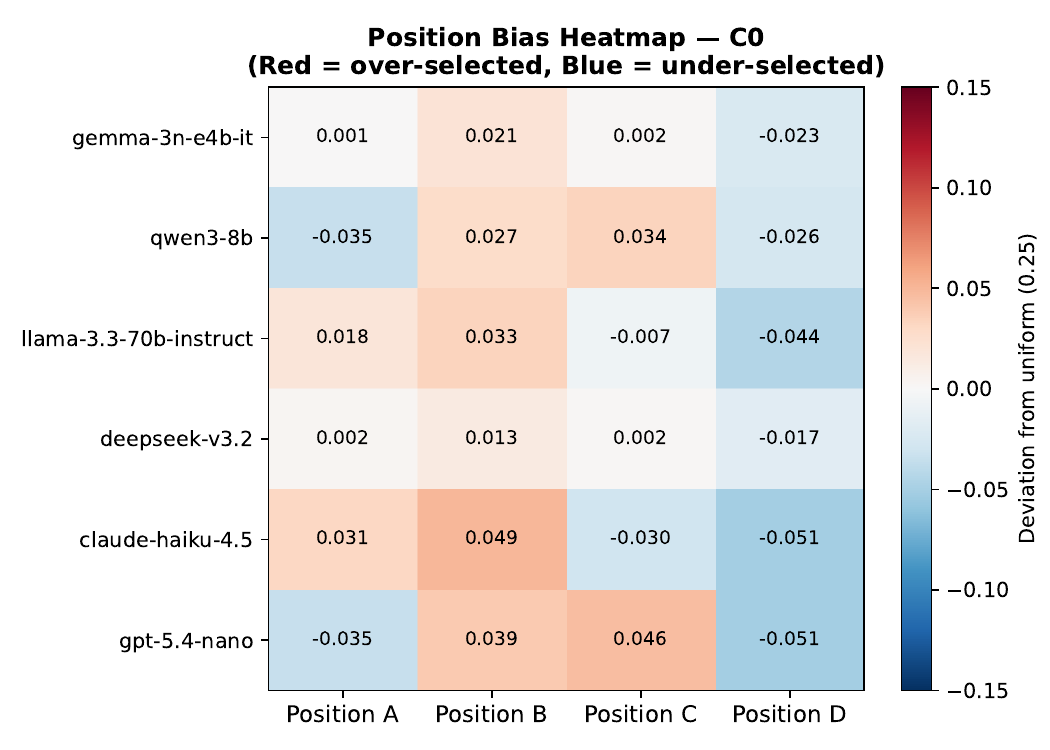}
\caption{Letter-position probabilities at C0 across all models after four option-order rotations. Values are close to the 0.25 uniform baseline, confirming that the cyclic rotation protocol effectively controls for letter-position preference.}
\label{fig:position_bias}
\end{figure}

\paragraph{Findings.}
After four option-order rotations, letter preferences are substantially controlled:
most probabilities fall within $\pm$0.05 of the 0.25 uniform baseline. Residual B-preference
visible in some models (e.g., Qwen C0: $P(B)=0.277$; LLaMA C0: $P(B)=0.283$) is
markedly reduced compared to single-rotation evaluations, confirming the value of the
cyclic rotation protocol. The small remaining deviations are unlikely to produce
systematic distortions in the cultural preference distributions after exposure normalisation.

% ---------------------------------------------------------------
% APPENDIX H: Extended Discussion
% ---------------------------------------------------------------
\FloatBarrier
\section{Extended Discussion}
\label{sec:appendix_discussion}

\FloatBarrier
\subsection{GPT-5.4 Nano Outlier Pattern}

GPT-5.4 Nano presents an instructive outlier pattern. Despite showing comparatively low
default concentration (KL $= 0.145$, Gini $= 0.349$, lower than four of the six models
at C0), it simultaneously exhibits the highest prior stickiness (PSI $= 0.504$, hardest
to steer at C2) and the highest JSD (0.018, most disrupted by fact injection). This
dissociation suggests that low default concentration does not guarantee good steerability:
the two properties are partially independent. A model may have a flatter, more
distributed default prior that is nonetheless firmly held and resistant to displacement.
This finding challenges the assumption that reducing default cultural bias (\eg through
data deduplication or diversity sampling) automatically improves cultural adaptability
under prompting. Future work should treat these as distinct axes of evaluation.

\FloatBarrier
\subsection{JSD Null Result Analysis}

The strikingly small JSD values ($< 0.02$ bits) challenge a common assumption in prompt
engineering: that providing explicit cultural facts for each option should shift the
model's selection distribution. Instead, the default prior remains almost entirely intact.
A plausible explanation is that the cultural facts are written in hedged, narrative
language that models do not treat as strong decision-relevant evidence, consistent with
findings that LLMs can be insensitive to factual context in MCQ settings
\citep{zheng2024llm}.

\FloatBarrier
\subsection{Primacy Bias Methodological Implications}

The primacy bias results add a further dimension: even when facts do shift selections,
part of that shift may be attributable to context-position effects rather than genuine
content processing, in line with the ``lost in the middle'' phenomenon documented by
\citet{liu2023lost}.

The primacy bias finding has methodological implications beyond this study. C3-style
evaluations that inject multiple facts sequentially without controlling context order
will systematically confound position effects with cultural preference effects. Our
$4\times4$ design provides a tractable correction that we recommend as a standard
control in future multi-fact injection evaluations. The fact that primacy bias severity
varies substantially across models (PPR range: 0.28 to 0.42) also suggests that
position sensitivity is a model-specific property worth characterising independently.

\FloatBarrier
\subsection{Cross-Culture PSI Pattern}

The cross-culture PSI pattern (high stickiness for Northern Nigeria, Ethiopia, Assam
and low stickiness for UK, US) aligns with the representational skew observed at C0
and likely reflects training data imbalance: models default to high-resource cultures and
resist steering away from them. This is consistent with BLEnD \citep{myung2024blend}
and CulturalBench \citep{chiu2025culturalbench}, which show similar patterns of model
underperformance on underrepresented cultures, and with GlobalOpinionQA
\citep{durmus2023global}, which reports that geographic prompting can fail to align
outputs for underrepresented populations.

\end{document}